%% file: main_arxiv.tex
\documentclass{article} 
\usepackage{iclr2027_conference,times}

\IfFileExists{math_commands.tex}{\input{math_commands.tex}}{}

\usepackage{amsmath,amssymb}
\usepackage{graphicx}
\usepackage{booktabs}   
\usepackage{multirow}  
\usepackage{array}
\usepackage{hyperref}
\usepackage{url}
\usepackage{subcaption}
\usepackage{dsfont}
\newcommand{\Ksel}{K}

\title{Look Where You Can: Active View Selection for CAD Reconstruction under Occlusion}

\author{Kartik Bali\textsuperscript{1}, Mahish K.~Guru\textsuperscript{2}, Yiderigun Borjigin\textsuperscript{3}, Alexandra Starostina\textsuperscript{1},\\
\textbf{Christian J.~Cyron}\textsuperscript{4}, \textbf{Roland Aydin}\textsuperscript{3} \\[4pt]
\textsuperscript{1}Helmholtz-Zentrum Hereon \quad
\textsuperscript{2}Leuphana Universit\"at L\"uneburg \\
\textsuperscript{3}Universit\"at des Saarlandes \quad
\textsuperscript{4}Technische Universit\"at Hamburg \\[2pt]
\texttt{kartik.bali@hereon.de}
}

\iclrfinalcopy 
\begin{document}

\maketitle
\lhead{Preprint}

\begin{abstract}
CAD reconstruction methods assume a luxury reality rarely grants:
unrestricted visual access to the object, photographed from any desired
angle. Real objects, however, are scene-embedded, bolted against walls,
wedged into corners, resting on floors, where the scene renders much of the
view sphere unreachable and the remaining views unequally informative. We
introduce \textbf{SightCAD}, a framework for parametric CAD reconstruction that treats view feasibility as a first-class constraint. In this work we consider objects from standard CAD benchmarks
embedded in realistic indoor scenes with physically derived visibility constraints over a discrete view sphere. A learned view selector
must choose $\Ksel$ feasible views for a vision--language model (VLM) that generates
executable CadQuery code, scored by geometric fidelity of the executed
solid. Because reward arrives only after discrete view selection, autoregressive
generation, and CAD-kernel execution, we propose a joint training paradigm in which the view selector and the CAD-generation VLM are trained together against this reward. The learned selection policy departs sharply from random, uniform, and coverage-greedy alternatives, outperforming surface-area maximization (SA-max) by up to $6.4$ Intersection-over-Union (IoU) points across budgets $\Ksel\in\{1,\dots,5\}$. The full system surpasses strong external baselines on scene-embedded, occluded multi-view renders of DeepCAD and Fusion360 objects ($+21$ and $+17$ effective-mIoU points over the best baseline, respectively), as well as on test-time domain-canonicalized real images from the industrial T-LESS benchmark and on both synthetic and real images from the MP6D industrial metal-parts benchmark, while producing the highest rate of executable programs of any method compared (invalid-code rate ${\leq}1.5\%$).
\end{abstract}

\section{Introduction}

\label{sec:intro}

Parametric computer-aided design (CAD) is the language in which engineers
specify physical structures, from bicycle pedals to airplanes to printed
circuit boards~\citep{wu2021deepcad,khan2024text2cad}. Its defining property is
that geometry is encoded not as a static mesh or point cloud but as an
\emph{editable, executable program}: a sketch-and-extrude sequence whose
numerical parameters can be inspected, re-solved, and re-manufactured.
Recovering such a program from observations of an existing object,
\emph{parametric CAD reconstruction}, is therefore the enabling step for
reverse-engineering, repair, and digitization, and a growing line of work
casts it as code generation from point clouds~\citep{rukhovich2025cad} or
images~\citep{kolodiazhnyi2025cadrille}.

The settings where this capability matters most are precisely those where it
is hardest to exercise: a part bolted to a factory wall, piping installed
along a ceiling, a fixture wedged into a cluttered interior. Existing
methods, however, from sketch-and-extrude generators~\citep{wu2021deepcad} to
vision--language models that emit CAD
code~\citep{rukhovich2025cad,kolodiazhnyi2025cadrille}, are trained and
evaluated on free-floating objects observed from unrestricted viewpoints.
This assumption is far from benign. A wall removes half the view sphere and a
corner still more, and the views that survive are far from equally
informative: one frames the part's defining feature, another reduces it to a
foreshortened smear, and a third shows nothing but a featureless box. A
system that cannot reason about \emph{where it may look}, and \emph{which of
the permitted views matter}, fails in exactly the deployments that motivate
the field.

In this work, we formalize this problem for CAD reconstruction. Placing an object in a realistic indoor scene renders part of the fixed view grid infeasible, so a method built to consume a fixed set of perspectives cannot exploit the views that remain available; a learned, budget-aware view selection strategy becomes necessary.

\begin{figure}[t]
\centering
\includegraphics[width=\linewidth]{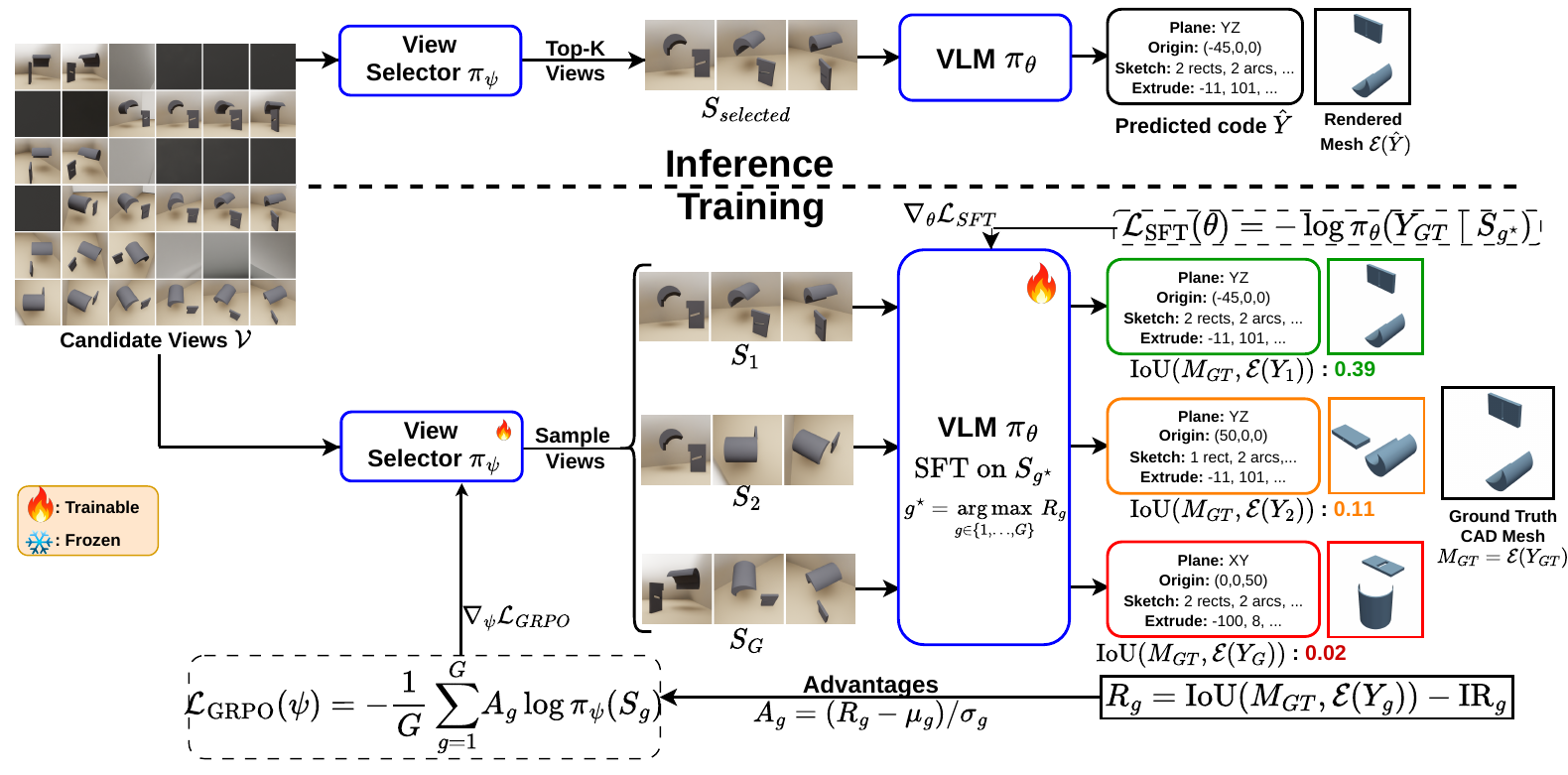}
\caption{\textbf{SightCAD} joint training paradigm: the view selector actively learns the VLM's view preferences via GRPO
while the VLM learns to understand and exploit view selections that maximize reward via SFT.
Here \(\psi\) and \(\theta\) are the trainable parameters of the view selector and the code-generation VLM, respectively; \(S\) is a subset of the calibrated views \(\mathcal{V}\); \(\mathrm{IR}\) is the invalid-code penalty ($-0.5$, Eq.\ref{eq:objective}); and \(M_g\) is the mesh computed by the execution operation \(\mathcal{E}\) (\(\mathrm{exec}\) in text) on code \(Y_g\).}
\label{fig:SightCAD_Stage1}
\end{figure}

Is learning necessary? Decades of next-best-view (NBV) heuristics
exist~\citep{connolly1985determination,scott2003view}, and greedy coverage
maximization, from viewpoint entropy~\citep{vazquez2002viewpoint} to
volumetric information gain~\citep{isler2016information}, is near-optimal for
coverage~\citep{nemhauser1978analysis}. But coverage is only a proxy for
reconstruction, and for program recovery it is a poor one: accuracy hinges on
a distortion-free readout of the program parameters from visual input. This
dependence on view geometry was recognized early in photogrammetric network
design~\citep{fraser1984network} and rediscovered by uncertainty-driven active
reconstruction~\citep{pan2022activenerf}. 
We show that a view selector trained end-to-end
with a CAD generator can learn a better view policy when rewarded for final reconstruction as it learns to trade coverage for parameter legibility.

Thus, our contributions are:
\begin{enumerate}
  \item \textbf{A new problem and dataset.} We formalize scene-embedded CAD
  reconstruction under physically derived view feasibility, and release a
  large-scale multi-view dataset in which objects from
  CAD-Recode~\citep{rukhovich2025cad} and
  Text2CAD~\citep{khan2024text2cad} are placed in
  SceneNet~\citep{handa2016understanding} rooms across three occlusion regimes
  (floor, wall, corner).
  \item \textbf{A joint selection--generation RL framework.} We propose \textbf{SightCAD}, a CAD reconstruction framework where we train a discrete view-subset policy jointly with an
  autoregressive CAD-code generator against an executable geometric reward,
  using shared rollouts so that credit is assigned separably to the choice
  of views and the choice of tokens.
  \item \textbf{An analysis of the learned perception strategy.} We show
  that the learned policy is interpretable and budget-dependent,  and give a mechanistic account of
  which views make which program parameters legible
  (Appendix~\ref{sec:analysis}).
  \item \textbf{State-of-the-art scene-embedded CAD reconstruction.} Our
  framework \textbf{SightCAD's} learned view selection policy outperforms SA-max selection at every budget
  $\Ksel\in\{1,\dots,5\}$, and the full system outperforms
  cadrille~\citep{kolodiazhnyi2025cadrille}, GenCAD~\citep{alam2024gencad}, and
  CAD-Recode$\leftarrow$TripoSR~\citep{tochilkin2024triposr} on DeepCAD,
  Fusion360, T-LESS~\citep{hodan2017t} and MP6D~\citep{chen2022mp6d} under matched scene occlusion, as well as real renders
  e.g., $68.5$ vs.\ $49.1$ mIoU on DeepCAD.
\end{enumerate}

\section{Related Work}
\label{sec:related}

\paragraph{Parametric CAD generation from visual input.}
DeepCAD~\citep{wu2021deepcad} introduced generative modeling of
sketch-and-extrude sequences, and subsequent work recast reconstruction as
\emph{code generation}: CAD-Recode translates point clouds into executable
programs with an LLM decoder~\citep{rukhovich2025cad}, cadrille extends
the recipe to multi-view images and fine-tunes its generator with online
RL~\citep{kolodiazhnyi2025cadrille}, and ReCAD likewise tunes a VLM with
geometric rewards~\citep{li2026recad}. A parallel line conditions generation on
a single image, either through diffusion and contrastive
priors~\citep{alam2024gencad} or by first lifting the image to a mesh and
then recovering the program~\citep{tochilkin2024triposr}. All of these
methods assume a free-space object and unconstrained viewpoints; none models
which views are available, and none selects among them. We inherit the
code-as-output formulation and the group-relative RL machinery, but make view
acquisition itself part of the learned policy.

\paragraph{Next-best-view planning and active reconstruction.}
View planning is a classical problem~\citep{connolly1985determination,
scott2003view}, and \emph{coverage} has been its dominant objective.
Viewpoint entropy scores a view by the entropy of its projected face
areas~\citep{vazquez2002viewpoint,polonsky2005s}, volumetric formulations
maximize newly observed
surface~\citep{isler2016information,delmerico2018comparison}, and
submodularity grants greedy selection $(1{-}1/e)$
near-optimality~\citep{nemhauser1978analysis,krause2014tractability}. Learned
variants inherit the same objective: NBV networks are supervised on
coverage~\citep{mendoza2020supervised,zeng2020pc}, and RL scanners reward it
directly~\citep{peralta2020next,chen2024gennbv}. A more recent wave
instead selects views by the uncertainty of an in-progress neural-field
reconstruction~\citep{pan2022activenerf,lee2022uncertainty,jin2023neu, jiang2023fisherrf}, and goal-oriented, frame-budgeted, and occlusion-aware
variants are beginning to
appear~\citep{zhu2026bayesian,song2026go,gao2026occamview}.
All of this work, however, outputs dense geometry, plans \emph{sequentially}
with the reconstruction in the loop, and optimizes objectives that are
independent of any downstream decoder. Our setting differs on multiple axes: the geometric output is a CadQuery program, views are chosen as a one-shot budgeted \emph{subset} by an amortized policy, feasibility is imposed by the scene, and the objective is the fidelity of the executed program.

\paragraph{Task-driven view selection.}
Closest to our mechanism, MVSelect trains an RL camera-selection module
jointly with a task network under a compute budget~\citep{hou2023learning},
and related policies choose views for active recognition and pose
estimation~\citep{jayaraman2016look,gartner2020deep} or for
manipulation under occlusion~\citep{bai2026learning}. In the CAD domain, MV-GEL ranks rendered views by the language-conditioned observability of a queried geometric entity for localization on meshes~\citep{bali2026mv}, though its views serve a discriminative grounding task and are scored independently rather than selected as a budgeted subset. These tasks are discriminative or control-oriented and provide dense feedback. On the theory side, our setting can be read as Bayesian experimental
design~\citep{lindley1956measure,foster2021deep} and mismatched
decoding~\citep{csiszar1995channel,lapidoth1996mismatched} applied to view selection; \citet{zhu2026bayesian} independently argue for task-specific NBV in a Bayesian framework, but target dense geometries and do not learn an
amortized policy.

\section{Dataset}
\label{sec:data}

In this section we describe our data creation pipeline, in which we embed CAD objects in generated indoor scenes, and illustrate our viewpoint sampling. We also describe our initial view selection policy, hereafter surface-area maximization (SA-max).

\paragraph{Source CAD models and scene placement.}
Ground-truth solids and their executable CadQuery programs come from
CAD-Recode~v1.5~\citep{rukhovich2025cad} (523k objects) and
Text2CAD~\citep{khan2024text2cad} (44k objects), spanning a broad range of sketch-and-extrude complexity. Each object is rescaled to a maximum extent of $0.8$--$1.0\,\mathrm{m}$ and given one placement in one of seven furnished SceneNet office-room variants, under one of three occlusion regimes, sampled uniformly: \textbf{floor} (free-standing, mild self-occlusion), \textbf{wall} (flush against one wall, one open half-space), and \textbf{corner} (wedged between two walls). Floor placements sample a collision-free floor position whose surrounding camera orbit is unobstructed; wall and corner placements draw from curated pools of 10 wall and 17 corner anchors (curation details in Appendix~\ref{sec:app:details}). As seen in Fig.~\ref{fig:dataset_scenes}, 
each object is placed at a random yaw in $[0,2\pi)$ with respect to the walls, with the camera grid rotated to compensate, so view indices stay in the object's canonical frame. 

\begin{figure}[t]
\centering
\includegraphics[width=\linewidth]{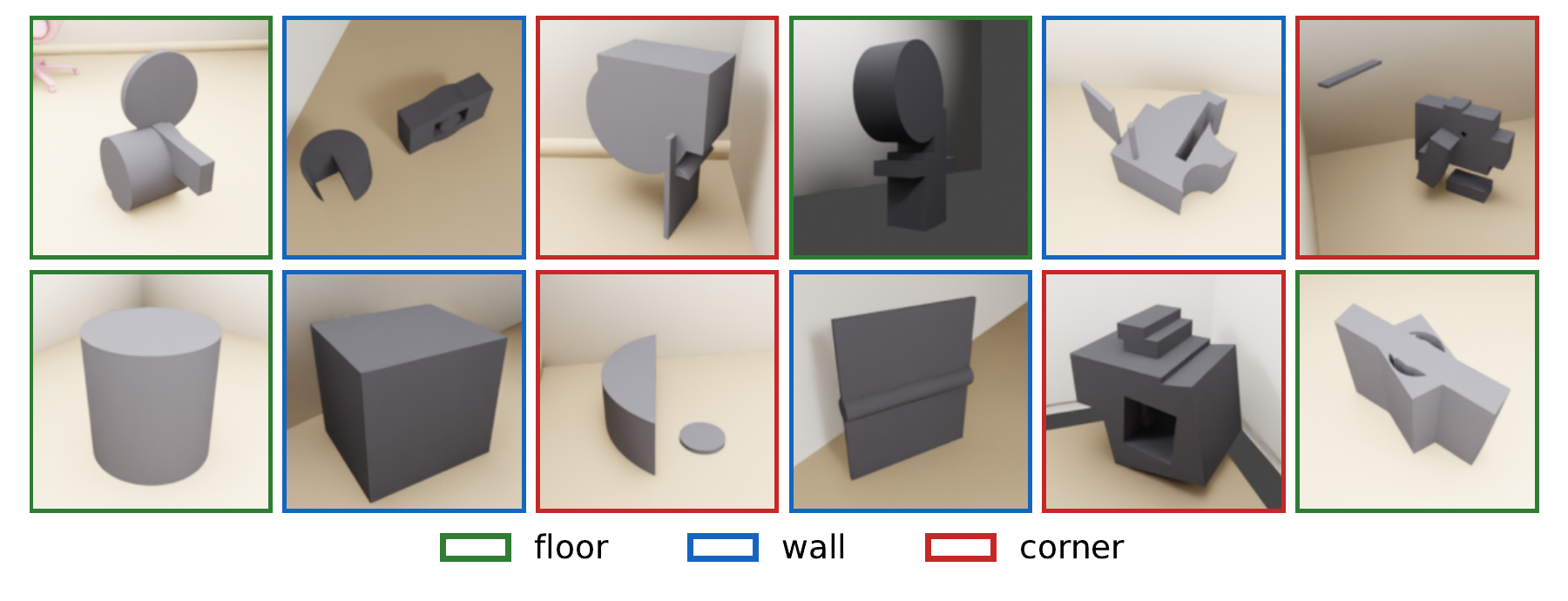}
\caption{\textbf{Scene-embedded CAD objects.} Objects from CAD-Recode and
Text2CAD are placed into furnished SceneNet office rooms under three occlusion
regimes: \emph{floor} (free-standing), \emph{wall} (one open half-space), and \emph{corner} (wedged between two walls). Placement determines which of the $N{=}36$ candidate views are feasible.}
\label{fig:dataset_scenes}
\end{figure}

\paragraph{Rendering and view feasibility.}
Scenes are rendered with BlenderProc~\citep{denningerblenderproc2} (Cycles)
at $256\times256$ under the room's own lighting plus local key/fill lights. Each
placement is captured from a fixed grid of $N{=}36$ viewpoints with three
elevation rings $\{0^{\circ},30^{\circ},60^{\circ}\}\times$ twelve azimuths
($30^{\circ}$ spacing) with each view tagged by its
canonical pose. This grid is the action space of the view selector. For each placement we sample $P{=}8000$ surface points (with area weights $a$)
and ray-cast from every camera to obtain a scene-aware visibility matrix
$V\in\{0,1\}^{N\times P}$, marking a point invisible under self- or scene
occlusion. The occlusion-aware coverage of view $i$ is the visible surface-area
fraction $c_i = \sum_p V_{i,p} a_p / \sum_p a_p$, and a view is \emph{feasible}
when $c_i > \epsilon$ ($\epsilon=0.01$), which masks out cameras inside or
behind walls. This yields our
SA-max policy: the ordering that repeatedly appends the view adding the most newly visible area (Appendix \ref{sec:app:selector-warmstart}).

\section{Method}
\label{sec:method}

\subsection{Problem formulation}
\label{sec:method:problem}

An object $O$ is embedded in a scene context $C$ and observed from a fixed
three-ring grid of candidate viewpoints (Sec.~\ref{sec:data}). A view is occluded if the camera
lies inside or behind scene geometry, or if the scene blocks sight of the
object. Each view provides a render $I_i$ with known pose $p_i$. Together these form the calibrated candidate set $\mathcal{V}=\{(I_i,p_i)\}_{i=1}^{N}$.
Given a budget $\Ksel$, a selection policy $\pi_{\psi}$ chooses a subset
$S \subset \{1,\dots,N\}$ of feasible views (Sec.~\ref{sec:data}),
$|S| = \Ksel$, and a generator produces a
CadQuery program $y \sim \pi_{\theta}(\cdot \mid \{I_i\}_{i \in S})$
whose execution yields a solid $\hat O = \mathrm{exec}(y)$; hereafter we
write $\pi_{\theta}(\cdot \mid S)$ for
$\pi_{\theta}(\cdot \mid \{I_i\}_{i \in S})$. The
joint objective is
\begin{equation}
\label{eq:objective}
\begin{split}
&\max_{\pi_{\psi},\, \pi_{\theta}}\;
\mathbb{E}_{S \sim \pi_{\psi}(\cdot \mid \mathcal{V})}\,
\mathbb{E}_{y \sim \pi_{\theta}(\cdot \mid S)}
\big[ R(y, O) \big],\\
&R(y,O) =
\begin{cases}
\mathrm{IoU}\big(\hat O,\, O\big) & y \text{ is executable and produces a non-empty solid},\\
-0.5 & \text{otherwise}
\end{cases}
\end{split}
\end{equation}
with volumetric IoU computed between voxelized solids. The reward is
observable only after discrete selection, autoregressive decoding, and CAD
kernel execution; no gradient path connects $R$ to either policy.

\subsection{Architecture}
\label{sec:method:arch}

\paragraph{View selector.}
The view selector $\pi_{\psi}$ maps the full calibrated view set $\mathcal{V}$ to per-view logits $z \in \mathbb{R}^{N}$. Each render is encoded by a frozen DINOv2
ViT-S/14 backbone~\citep{oquab2023dinov2}; pose is injected by sinusoidal
azimuth/elevation encodings; a cross-view transformer exchanges information
across the $N$ view candidates so that logits reflect \emph{joint} evidence for per-view quality (for further details on the architecture, refer to Appendix \ref{sec:app:selector-arch}). During training,
subsets are drawn without replacement by sequential masked sampling from the
softmax over the logits $z$ (Eq.\ref{eq:pl}; \citealp{singh2019policy});
at inference the deterministic top-$\Ksel$ views are taken.
\begin{equation}
\label{eq:pl}
i_k \;\sim\; P(i \mid i_1,\dots,i_{k-1})
\;=\;
\frac{\exp z_i \,\cdot\, \mathds{1}\!\left[i \notin \{i_1,\dots,i_{k-1}\}\right]}
     {\sum_{j \notin \{i_1,\dots,i_{k-1}\}} \exp z_j},
\qquad k = 1,\dots,\Ksel
\end{equation}

Summing the per-step log-probabilities of Eq.\ref{eq:pl} gives the
log-probability the policy assigns to the whole ordered draw $S=(i_1,\dots,i_{\Ksel})$,
\begin{equation}
\label{eq:subset-logprob}
\log \pi_{\psi}(S)
\;=\;
\sum_{k=1}^{\Ksel} \log
\frac{\exp z_{i_k}}
     {\sum_{j \notin \{i_1,\dots,i_{k-1}\}} \exp z_j}.
\end{equation}

\paragraph{Generator.}
The generator $\pi_{\theta}$ is a Qwen2-VL vision--language
model~\citep{wang2024qwen2} ($\sim$2B parameters) initialized from the
multi-view image-to-CadQuery checkpoint of cadrille~\citep{kolodiazhnyi2025cadrille},
which supplies a strong sketch-and-extrude code prior. It takes the $\Ksel$
views selected by $\pi_{\psi}$ and concatenates them in a horizontal strip as input: each render is tokenized into visual tokens by the vision encoder and interleaved, in ascending view-index order, with a short
textual pose tags ($\texttt{e}\langle\text{elev}\rangle\texttt{\_a}\langle\text{azim}\rangle$) for each view in the strip, derived from the camera's position on the view grid. Conditioned on this multimodal prompt, $\pi_{\theta}$ autoregressively emits a program $y$.

\begin{figure}[t]
    \centering
    \includegraphics[width=\linewidth]{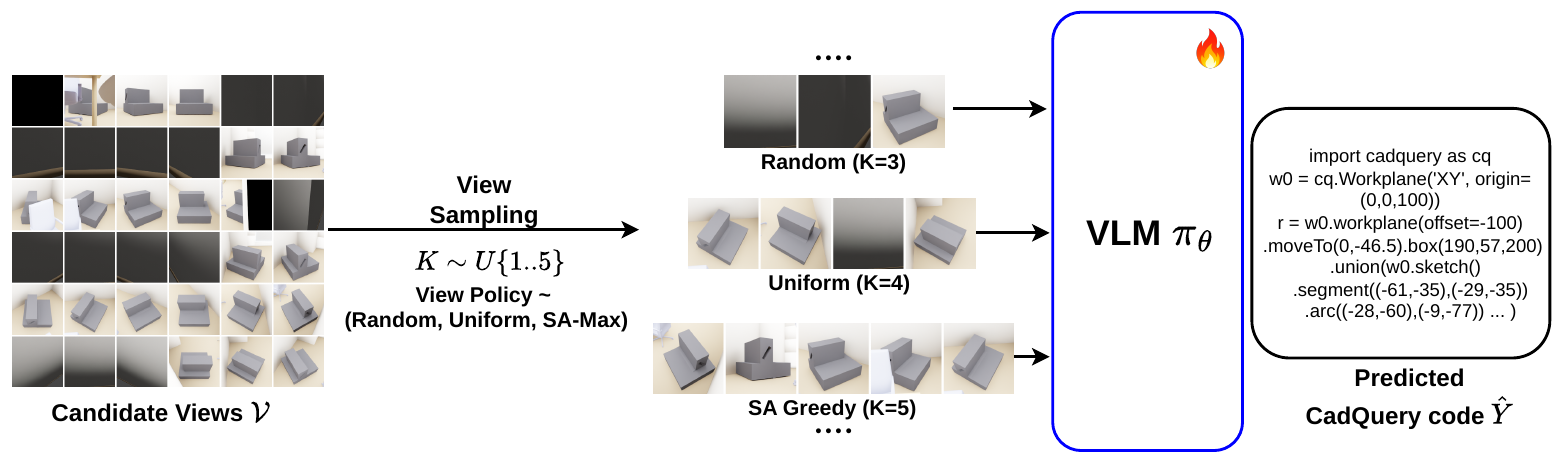}
    \caption{\textbf{SightCAD} generator warm-starting on variable-$K$ multi-view subsets: candidate
    views are sampled under three view policies (random, uniform, SA-max)
    with $K\sim\mathrm{U}\{1,\dots,5\}$}
    \label{fig:vlm-variablek}
\end{figure}

\subsection{Training}
\label{sec:method:training}
We divide training into three phases: \emph{Network Warm-Starting}, where we
prime both the view selector and the generator VLM for stable downstream RL;
\emph{Joint Training}, where the view selector is trained with Group Relative
Policy Optimization (GRPO)~\citep{shao2024deepseekmath} while the generator VLM
is concurrently fine-tuned with the negative log-likelihood (NLL) of the ground-truth CadQuery code; and
\emph{Generator RL Tuning}, where the view selector is frozen and the VLM is
trained with GRPO on its own high-temperature samples, rewarded against the
ground-truth geometry. Each stage runs for one epoch over the entire scene-embedded CAD dataset (Sec.~\ref{sec:data}).

\paragraph{Network Warm-Starting.}
\label{sec:method:training:network_warm_starting}
Using an untrained VLM as the reward model leads to unstable RL training due
to uninformative initial rewards. We therefore prime the CadQuery generator to
handle variable viewpoints and view counts: it is fully fine-tuned with token-level
NLL on ground-truth programs, conditioned on view sets drawn from a curriculum
mixing the SA-max, random, and uniformly spaced view policies,
with $K \sim \mathrm{U}\{1,\dots,5\}$ (Fig.~\ref{fig:vlm-variablek}). The uniformly spaced
policy is a naive, scene-agnostic orbit scan
(Appendix~\ref{sec:app:details}). This mixture
prevents the generator from overfitting to any fixed view rig and prepares it
to exploit whatever views the selector later provides. We likewise initialize
the view selector by distilling an already viable SA-max
strategy (Appendix~\ref{sec:app:selector-warmstart}, Fig.~\ref{fig:sa-max}) that naturally prioritizes object visibility and filters out occluded views.

\paragraph{Joint Training: View Selector RL + Generator SFT.}
After initialization, the view selector policy $\pi_{\psi}$ is
optimized with GRPO adapted to view subsets (Fig.~\ref{fig:SightCAD_Stage1}). For each object, a group of $G$
subsets $S_1,\dots,S_G \sim \pi_{\psi}$ is sampled via
Eq.\ref{eq:pl}. The warm-started generator $\pi_{\theta}$ decodes each subset
greedily, and the rewards $R_g$ from Eq.\ref{eq:objective} are converted to
group-relative advantages, which weight the log-probabilities of the sampled
subsets to form the selector loss (Eq.\ref{eq:grpo}):
\begin{equation}
\label{eq:grpo}
A_g = \frac{R_g - \operatorname{mean}(R_{1:G})}
           {\operatorname{std}(R_{1:G}) + \varepsilon},
\qquad
\mathcal{L}_{\mathrm{GRPO}}(\psi) = -\frac{1}{G} \sum_{g=1}^{G} A_g \log
\pi_{\psi}(S_g).
\end{equation}
Concurrently, the generator $\pi_{\theta}$ is trained with the negative
log-likelihood of the ground-truth code $Y_{\mathrm{GT}}$ conditioned on the
highest-reward view set in the group:
\begin{equation}
\label{eq:sft}
\mathcal{L}_{\mathrm{SFT}}(\theta) = -\log
\pi_{\theta}\!\left(Y_{\mathrm{GT}} \mid S_{g^\star}\right),
\qquad
g^{\star}=\underset{g\in\{1,\dots,G\}}{\arg\max}\; R_{g}.
\end{equation}
Because all subsets in a group share the same object and the generator decodes
greedily, reward differences within a group are largely attributable to the selected views. The SFT update is additionally gated by ground-truth view
visibility, so the generator is never trained to hallucinate unseen geometry.
We sample $G{=}4$ subsets per object across $4$ GPUs in Distributed Data
Parallel~\citep{li2020pytorch}; optimizer and visibility-gate hyperparameters
are listed in Appendix~\ref{sec:app:details} (Table~\ref{tab:hparams}).

\begin{figure}[t]
    \centering
    \includegraphics[width=\linewidth]{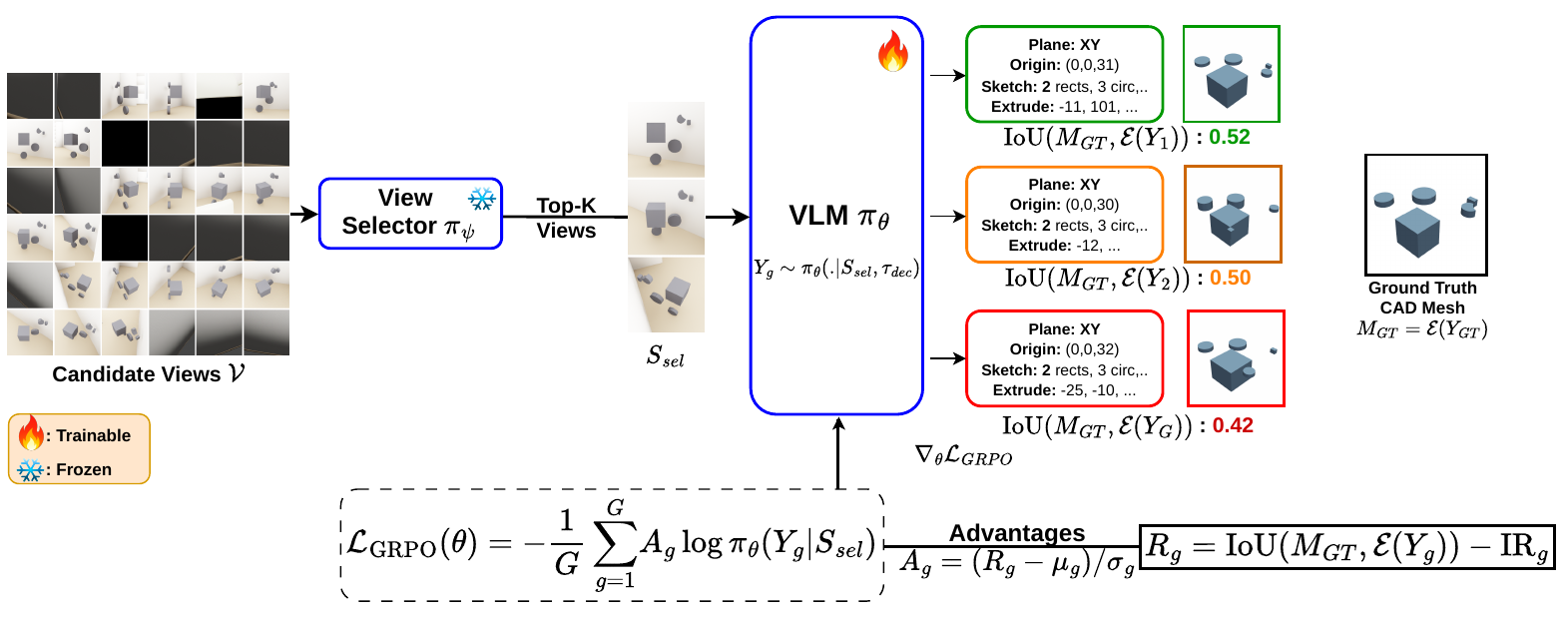}
    \caption{\textbf{SightCAD} generator RL tuning: the view selector trained in the joint stage is kept frozen while the VLM further learns to understand views and express code better via GRPO on high-temperature code rollouts. All mathematical expressions are as per Fig. \ref{fig:SightCAD_Stage1}. KL regularization toward the SFT reference ($\beta\,\widehat{\mathbb{D}}_{\mathrm{KL}}[\pi_\theta \,\|\, \pi_{\mathrm{ref}}]$, Eq.\ref{eq:grpo_gen}) omitted for clarity.}
    \label{fig:SightCAD_stage2}
\end{figure}

\paragraph{Generator RL Tuning.}
Once the view selector converges, it is frozen and the \emph{joint-trained} generator
$\pi_{\theta}$ is further optimized with standard GRPO over $G$ decodes
$Y_1,\dots,Y_G$ sampled at high temperature ($\tau{=}0.9$), as shown in Fig.~\ref{fig:SightCAD_stage2}, conditioned on
the frozen selector's views $S^{\star}$ and scored with the same executable
IoU reward. For training stability we regularize the policy toward the
initialization checkpoint $\pi_{\mathrm{ref}}$ with a per-token KL penalty:
\begin{equation}
\label{eq:grpo_gen}
\mathcal{L}_{\mathrm{GRPO}}(\theta) = -\frac{1}{G} \sum_{g=1}^{G}
\frac{1}{|Y_g|} \sum_{t=1}^{|Y_g|}
\Big[ A_g \log \pi_{\theta}\!\left(y_{g,t} \mid y_{g,<t},\, S^{\star}\right)
\;-\; \beta\, \widehat{\mathbb{D}}_{\mathrm{KL},t}\!\left[\pi_{\theta}
\,\|\, \pi_{\mathrm{ref}}\right] \Big],
\end{equation}
where the advantages $A_g$ are computed as in Eq.\ref{eq:grpo} and the KL
term uses the unbiased non-negative estimator
of~\citet{shao2024deepseekmath}:
\begin{equation}
\label{eq:kl_k3}
\widehat{\mathbb{D}}_{\mathrm{KL},t}\!\left[\pi_{\theta} \,\|\,
\pi_{\mathrm{ref}}\right]
= \frac{\pi_{\mathrm{ref}}(y_{g,t} \mid y_{g,<t}, S^{\star})}
       {\pi_{\theta}(y_{g,t} \mid y_{g,<t}, S^{\star})}
- \log \frac{\pi_{\mathrm{ref}}(y_{g,t} \mid y_{g,<t}, S^{\star})}
            {\pi_{\theta}(y_{g,t} \mid y_{g,<t}, S^{\star})}
- 1 .
\end{equation}
For $\widehat{\mathbb{D}}_{\mathrm{KL},t}\!\left[\pi_{\theta} \,\|\,
\pi_{\mathrm{ref}}\right]$ we choose $\pi_{\mathrm{ref}}$ to be the final SFT
weights of the \emph{joint-trained} generator; the remaining hyperparameters
are listed in Appendix~\ref{sec:app:details} (Table~\ref{tab:hparams}).
Decoupling the two RL phases assigns GRPO credit to the correct module in each: a single combined RL objective over both $\pi_{\psi}$ and $\pi_{\theta}$ would confound whether reward variation stems from the views or from the decode.

\section{Evaluation}
\label{sec:experiments}

\paragraph{Data and metrics.}
We evaluate our framework on a held-out test split of the CAD-Recode and Text2CAD datasets comprising $\sim$4k CAD objects for view policy comparison, and compare against external baselines on the DeepCAD-test and Fusion360 datasets, with object placements spanning floor/wall/corner occlusion regimes. Placements are randomly sampled and frozen, thus every method and view policy is evaluated on identical placements and occlusion patterns. For a more realistic setting, we further use two multi-view industrial CAD datasets: T-LESS~\citep{hodan2017t}, with both synthetic re-renders of the CAD models ($n{=}30$) and real captured images ($n{=}138$ multi-view sets), and the metal-parts dataset MP6D~\citep{chen2022mp6d} ($n{=}20$ synthetic, $n{=}200$ real multi-view sets extracted from video frames). We report invalidity ratio (IR\%: generations that fail to execute or produce an empty solid), median Chamfer distance ($\times 10^{3}$, computed on valid samples only), and mean volumetric IoU (mIoU\%, over valid samples). We additionally report an \emph{effective} variant (mIoU\textsubscript{eff}) that scores every sample, assigning zero IoU to invalid generations.
These metrics are computed under two scoring protocols i.e \emph{view-policy comparison} and \emph{cross-method comparison}, matched to what each can assume. The view-policy comparisons (Table~\ref{tab:schemes-main}) share our canonical frame and metric scale by construction and thus absolute-size and placement errors are penalized as genuine program errors. The cross-method comparisons (Tables \ref{tab:external}, \ref{tab:tless}, \ref{tab:mp6d}) involve baselines whose outputs carry arbitrary units. We therefore adopt a scale-invariant protocol that normalizes each predicted and ground-truth mesh independently to the unit cube before computing IoU and Chamfer distance, so that no method is penalized for its unit rather than for its
geometry.

\paragraph{Evaluation structure and baselines.}
We organize the evaluation into three comparisons.
\emph{(i) Co-adapted generator}: each policy is paired with the generator co-trained on its own views (Sec.~\ref{sec:method:training}), reported at both stages, \emph{SFT} and \emph{+GRPO}, with a separate pair per budget $\Ksel\in\{1,\dots,5\}$, contrasting the learned policy against SA-max and its curvature-weighted variant fw-SA (Appendix~\ref{sec:app:details}).
\emph{(ii) Policy-neutral generator}: the single warm-started generator (Sec.~\ref{sec:method:training:network_warm_starting}) decodes random, uniform, SA-max, fw-SA, and \emph{learned} selections, isolating view quality from co-adaptation. For both \emph{(i)} and \emph{(ii)} we use our test split of $\sim$4k CAD objects.
\emph{(iii) External CAD baselines}: \textbf{SightCAD} ($\Ksel{=}4$, SFT and +GRPO) against \emph{cadrille}~\citep{kolodiazhnyi2025cadrille}, \emph{CAD-Coder}~\citep{doris2026cad}, \emph{CAD-Fit}~\citep{nehme2026cadfit}, \emph{CAD-Recode$\leftarrow$TripoSR}~\citep{tochilkin2024triposr,rukhovich2025cad}, and \emph{GenCAD}~\citep{alam2024gencad}, each on its native input convention (Appendix~\ref{sec:app:qual}).

\begin{table*}[t]
\centering
\caption{\textbf{What view selection is worth, and what co-adaptation adds.}
\emph{Left} (\textbf{a}): each policy is paired with the generator trained
alongside it, at both stages (SFT, +GRPO).
\emph{Right} (\textbf{b}): a single policy-neutral \emph{warm-started} generator
decodes the views chosen by five selection policies
fw-SA is a curvature-saliency-reweighted variant of SA-max
(Appendix~\ref{sec:app:details}). Scene-embedded test split, mIoU\textsubscript{eff} (\%); standard
errors are $0.46$--$0.53$ throughout ($n{=}4000$ objects).}
\label{tab:schemes-main}
\small
\setlength{\tabcolsep}{4.5pt}
\resizebox{\textwidth}{!}{%
\begin{tabular}{c cc cc cc ccccc}
\toprule
& \multicolumn{2}{c}{SA-max (\textbf{a})}
& \multicolumn{2}{c}{fw-SA (\textbf{a})}
& \multicolumn{2}{c}{Learned (\textbf{a})}
& \multicolumn{5}{c}{Policy-neutral generator (\textbf{b})} \\
\cmidrule(lr){2-3}\cmidrule(lr){4-5}\cmidrule(lr){6-7}\cmidrule(lr){8-12}
$\Ksel$ & SFT & +GRPO & SFT & +GRPO & SFT & +GRPO
        & Random & Uniform & SA-max & fw-SA & Learned \\
\midrule
1 & 45.80 & 53.05 & 44.15 & 50.98 & \textbf{49.97} & \textbf{55.86} & 24.36 & 23.79 & 40.90 & 39.23 & \textbf{44.05} \\
2 & 47.36 & 54.28 & 46.72 & 53.75 & \textbf{50.68} & \textbf{56.95} & 33.11 & 36.40 & 44.44 & 44.32 & \textbf{46.04} \\
3 & 47.75 & 54.72 & 47.66 & 54.40 & \textbf{51.69} & \textbf{58.51} & 36.89 & 40.57 & 46.02 & 46.11 & \textbf{47.46} \\
4 & 48.15 & 55.16 & 48.26 & 54.99 & \textbf{52.90} & \textbf{60.52} & 38.82 & 42.68 & 46.30 & 46.59 & \textbf{47.81} \\
5 & 48.55 & 55.40 & 48.89 & 55.29 & \textbf{53.88} & \textbf{61.79} & 39.85 & 43.81 & 46.01 & 46.68 & \textbf{48.86} \\
\bottomrule
\end{tabular}%
}\\[2pt]
\end{table*}

\paragraph{View Selector Co-adapted Generator.}
Table~\ref{tab:schemes-main}a pairs each policy with its co-trained
generator. IoU increases monotonically with the budget for both policies and
both training stages, and the learned selector outperforms SA-max at
every $\Ksel$, by $3.3$ to $5.3$ mIoU after joint training (SFT) and by up to $+6.4$
after generator RL (+GRPO). This gap holds within every occlusion regime (Appendix Table~\ref{tab:sig-regime}). Generator RL training improves both policies over their SFT stage, and the improvement is larger for the learned views. The fw-SA columns test whether a part-aware coverage objective closes this gap: despite the transfer disadvantage of decoding through the SA-max--co-adapted generator, fw-SA matches SA-max within $0.7$ mIoU at every budget $\Ksel\geq2$ (at $\Ksel{=}3$: $47.66$ vs.\ $47.75$ SFT and $54.40$ vs.\ $54.72$ +GRPO; see Appendix \ref{sec:app:insitu4k}), so curvature-weighted coverage is interchangeable with plain coverage yet remains far below the learned policy.

\paragraph{Fixed Policy Neutral Generator.}
Table~\ref{tab:schemes-main}b isolates view quality by decoding all five policies with
the single policy-neutral generator (Sec.~\ref{sec:method:training:network_warm_starting}). The ordering
learned $>$ fw-SA $\approx$ SA-max $>$ uniform $>$ random holds at every budget except
$\Ksel{=}1$, where uniform trails random: the learned views are on
average more informative for CadQuery code prediction, and the
learned-over-SA-max gap is significant at every budget. The gaps separate the
value of selection into two components: occlusion-aware coverage selection
contributes the most (SA-max over random), and image-based learned selection
adds a further $+1.4$ to $+3.2$ over SA-max at every budget. Under this fully
policy-neutral decode, fw-SA is statistically indistinguishable from SA-max at
$\Ksel\in\{2,3,4\}$, worse at $\Ksel{=}1$, and $+0.7$ at $\Ksel{=}5$, while the learned policy stays significantly ahead of both at every
budget (for a mechanistic analysis, refer to Appendix \ref{sec:mechanism}).

\begin{table}[t]
\centering
\caption{\textbf{DeepCAD-test and Fusion360-test ($n{=}100$ each).}}
\label{tab:external}
\setlength{\tabcolsep}{4pt}
\small
\resizebox{\linewidth}{!}{%
\begin{tabular}{@{}l rrrr rrrr@{}}
\toprule
& \multicolumn{4}{c}{DeepCAD} & \multicolumn{4}{c}{Fusion360} \\
\cmidrule(lr){2-5}\cmidrule(lr){6-9}
Method & IR\% $\downarrow$ & medCD $\downarrow$ & mIoU\% $\uparrow$ & mIoU\textsubscript{eff} $\uparrow$
       & IR\% $\downarrow$ & medCD $\downarrow$ & mIoU\% $\uparrow$ & mIoU\textsubscript{eff} $\uparrow$ \\
\midrule
Ours ($\Ksel{=}4$, +GRPO)     & \textbf{0.00} & \textbf{5.36} & \textbf{68.49} & \textbf{68.49} & \textbf{1.00} & \textbf{10.42} & \textbf{53.56} & \textbf{53.03} \\
Ours ($\Ksel{=}4$, SFT)       & 0.00  & 6.83  & 62.18 & 62.18 & 1.00  & 11.96 & 49.86 & 49.36 \\
cadrille                       & 4.00  & 18.30 & 49.05 & 47.09 & 5.00  & 33.62 & 37.73 & 35.85 \\
CAD-Fit                        & 6.00  & 21.07 & 47.08 & 44.26 & 17.00 & 26.46 & 40.32 & 33.47 \\
CAD-Coder                      & 2.00  & 37.10 & 38.29 & 37.52 & 3.00  & 66.61 & 28.56 & 27.71 \\
CAD-Recode$\leftarrow$TripoSR  & 16.00 & 76.92 & 18.84 & 15.83 & 15.00 & 80.07 & 18.10 & 15.38 \\
GenCAD                         & 58.00 & 91.04 & 18.21 & 7.65  & 77.00 & 46.77 & 16.43 & 3.78  \\
\bottomrule
\end{tabular}%
}
\end{table}

\begin{table*}[t]
\centering
\caption{\textbf{T-LESS industrial objects ($n{=}138$ real, $n{=}30$ synthetic)}.}
\label{tab:tless}
\setlength{\tabcolsep}{4pt}
\footnotesize
\resizebox{\textwidth}{!}{%
\begin{tabular}{@{}l rrrr rrrr rrrr@{}}
\toprule
& \multicolumn{4}{c}{\emph{Synth} ($n{=}30$)} & \multicolumn{4}{c}{\emph{Real} ($n{=}138$)} & \multicolumn{4}{c}{\emph{Canon} ($n{=}138$)} \\
\cmidrule(lr){2-5}\cmidrule(lr){6-9}\cmidrule(lr){10-13}
Method & IR\% & medCD & mIoU\% & mIoU\textsubscript{eff}
       & IR\% & medCD & mIoU\% & mIoU\textsubscript{eff}
       & IR\% & medCD & mIoU\% & mIoU\textsubscript{eff} \\
\midrule
Ours ($\Ksel{=}4$, +GRPO)     & \textbf{0.00} & 18.79 & 54.02 & \textbf{54.02} & \textbf{1.45} & 111.59 & 8.38  & 8.26          & \textbf{0.00} & \textbf{43.56} & \textbf{32.48} & \textbf{32.48} \\
Ours ($\Ksel{=}4$, SFT)       & 3.33  & 24.47 & 46.76 & 45.20 & 4.35  & 101.41 & 9.68  & 9.25          & 11.59 & 71.36 & 19.83 & 17.53 \\
cadrille          & 10.00 & 36.45 & 34.92 & 31.43 & 24.64 & 104.27 & 13.39 & 10.09         & 27.54 & 81.21 & 15.11 & 10.95 \\
CAD-Fit                       & 13.33 & \textbf{12.53} & \textbf{57.73} & 50.03 & 57.97 & \textbf{45.54} & \textbf{34.33} & 14.43 & 39.86 & 47.54 & 29.86 & 17.96 \\
CAD-Coder         & 6.67  & 84.22 & 17.94 & 16.74 & 23.91 & 94.36  & 15.58 & 11.86         & 2.90 & 58.36 & 27.11 & 26.32 \\
CAD-Recode$\leftarrow$TripoSR & 6.67  & 40.23 & 34.83 & 32.51 & 17.39 & 47.65  & 31.06 & \textbf{25.66} & 13.04 & 65.47 & 23.02 & 20.01 \\
GenCAD            & 96.67 & 73.66 & 14.24 & 0.47  & 95.65 & 93.62  & 11.48 & 0.50          & 51.45 & 86.42 & 17.74 & 8.61  \\
\bottomrule
\end{tabular}%
}
\end{table*}

\begin{table*}[t]
\centering
\caption{\textbf{MP6D industrial objects ($n{=}200$ real, $n{=}20$ synthetic).}}
\label{tab:mp6d}
\setlength{\tabcolsep}{4pt}
\footnotesize
\resizebox{\textwidth}{!}{%
\begin{tabular}{@{}l rrrr rrrr@{}}
\toprule
& \multicolumn{4}{c}{\emph{Synth} ($n{=}20$)} & \multicolumn{4}{c}{\emph{Real} ($n{=}200$)} \\
\cmidrule(lr){2-5}\cmidrule(lr){6-9}
Method & IR\% & medCD & mIoU\% & mIoU\textsubscript{eff}
       & IR\% & medCD & mIoU\% & mIoU\textsubscript{eff} \\
\midrule
Ours ($\Ksel{=}4$, +GRPO)
& \textbf{0.00} & \textbf{20.55} & \textbf{48.31} & \textbf{48.31}
& \textbf{0.50} & \textbf{57.91} & \textbf{25.71} & \textbf{25.58} \\

Ours ($\Ksel{=}4$, SFT)
& 10.00 & 31.22 & 40.81 & 36.73
& 4.00 & 72.09 & 19.37 & 18.59 \\

cadrille
& 10.00 & 26.91 & 43.03 & 38.73
& 30.50 & 77.94 & 18.44 & 12.82 \\

CAD-Fit
& 20.00 & 29.51 & 42.39 & 33.91
& 21.50 & 71.03 & 21.85 & 17.16 \\

CAD-Coder
& 15.00 & 49.47 & 27.51 & 23.38
& 7.50 & 66.28 & 21.45 & 19.84 \\

CAD-Recode$\leftarrow$TripoSR
& 15.00 & 42.30 & 30.91 & 26.27
& 14.50 & 68.81 & 18.65 & 15.95 \\

GenCAD
& 80.00 & 112.53 & 9.19 & 1.84
& 65.00 & 103.74 & 7.55 & 2.64 \\

\bottomrule
\end{tabular}%
}
\end{table*}

\paragraph{Comparison with Other CAD Baselines.} Tables~\ref{tab:external},~\ref{tab:tless}, and~\ref{tab:mp6d} report our framework evaluated on external datasets against existing CAD baselines. Every baseline is evaluated on its \emph{own} native input format, and the scene contributes \emph{only occlusion}: where a native viewpoint is occluded by the object's scene-embedded placement, the corresponding view is replaced by a blank frame (per-baseline conventions in Appendix~\ref{sec:app:qual}).
\textbf{SightCAD} ($\Ksel{=}4$, +GRPO) achieves the best
mIoU\textsubscript{eff} on the scene-embedded benchmarks (DeepCAD-test,
Fusion360, T-LESS synth, MP6D synth) and on canonicalized real captures (T-LESS canon and MP6D real), and emits the most
executable CAD: its invalid rate is $1.45\%$ on real photographs and $0.00\%$
after canonicalization. The T-LESS \emph{real} evaluation feeds real photographs from the wild
directly to each method. The \emph{canon} evaluation applies a
model-agnostic, test-time-only domain match (SAM
segmentation~\citep{kirillov2023segment}, exposure normalization, and
compositing onto each baseline's own native plate, without retraining or ground
truth), which places every CAD viewpoint in its most favorable
in-distribution image setting. On T-LESS real images our approach trails some baselines due to an image-distribution gap that canonicalization removes. Unlike the CAD-benchmark evaluations, where
the pose used to author the ground-truth program is known, neither T-LESS nor MP6D provides a
canonical frame, and their real captures lie on scattered camera poses rather than on our discrete
view grid. For our method only, we therefore apply a fixed two-step protocol before decoding: a
learned alignment step (\emph{align-net}) recovers the in-plane object orientation and re-anchors
all view azimuths to it, and the available views are then snapped onto the selector's training-time
view lattice (one view per slot, blank tiles for unfilled slots) so that the frozen selector chooses
its $\Ksel$ views exactly as at training time. Both steps are applied identically to the synthetic,
raw, and canonicalized conditions of each benchmark (details in
Appendices~\ref{sec:app:alignnet} and~\ref{sec:app:realview-selection}).
On MP6D (Table~\ref{tab:mp6d}), whose captures are acquired on a plain background with no embedding scene, the raw--canon domain match required for T-LESS does not apply, so we report the synthetic and real conditions only; there \textbf{SightCAD} ($\Ksel{=}4$, +GRPO) leads every baseline on \emph{both}, corroborating that the T-LESS real deficit is a background-domain artifact rather than a multi-view reconstruction failure. For comparison of baselines on the internal  $\sim$4k test split of our dataset, please refer to Table \ref{tab:insitu4k-baselines} in the Appendix.

\vspace{-8pt}
\paragraph{Learned View Policy Analysis.}
View selection changes strongly with the budget $\Ksel$: a single mid-elevation
canonical view at $\Ksel{=}1$, a narrow stereo wedge that pins the defining
feature's depth at $\Ksel{=}2$, and from $\Ksel{=}3$ an orthographic
front/side/plan drawing triad, recovering the projection conventions of a
mechanical drawing from reward alone. We also probe mechanistically, locating each view's visual tokens in the horizontal strip of the $\Ksel$ views. At every decoder layer, we ablate the face-on (views angled at $\leq15^\circ$ to the part's sketch normal) and in-plane (views angled at $\geq75^\circ$) view tokens by replacing them with the mean of the other views' tokens, keeping hidden states in-distribution. We then teacher-force the ground-truth CadQuery program and measure the resulting $\Delta \mathrm{NLL}$ on specific token categories, extrude-depth digits and sketch-coordinate digits. This reveals that the learned policy supplies more of the in-plane views that carry the signal for accurate extrude-depth inference than any other view policy considered in this work (Fig.~\ref{fig:mechanism-summary}). The detailed
analysis is in Appendix~\ref{sec:analysis}.


\vspace{-2pt}
\section{Conclusion}
We introduced scene-embedded CAD reconstruction under physically derived view feasibility and \textbf{SightCAD}, a framework that treats view acquisition as a learned, budgeted decision. \textbf{SightCAD} jointly trains a discrete view-subset selector and an autoregressive CadQuery generator against the volumetric IoU of the executed solid using shared-rollout GRPO, separating credit for the choice of views from credit for the choice of tokens. The learned selector outperforms SA policies at every budget, and the full system leads strong external baselines under matched scene occlusion while emitting executable code far more reliably. The learned policy partly recovers engineering-drawing projection conventions, prioritizing views that make program parameters, especially extrude depth, legible.

\vspace{-8pt}
\paragraph{Limitations and future work.} \textbf{SightCAD} selects from a fixed discrete view grid and assumes known camera parameters, object distance, and canonical anchor (compensable only partly at test time). Continuous camera control, relaxing the anchor assumption, and closing the raw-photograph gap are natural next steps. Our interpretability findings are moreover tied to the specific Qwen2-VL generator we train and may not transfer to other VLM backbones or tokenizations.

\subsection*{AI use statement}

In this work, we used AI coding assistants to help with minor  implementations of the
training, rendering, and evaluation code. We have not used generative AI tools for creating CAD
data: all training and evaluation data derive from existing public assets
(CAD-Recode, Text2CAD, SceneNet, DeepCAD, Fusion360, T-LESS, MP6D) processed by our
deterministic placement, rendering, and visibility pipeline. We additionally
used generative AI tools for light editing of the manuscript text (grammar and
phrasing). All AI-assisted work was reviewed by the authors: AI-generated code
was verified and tested for correctness by the authors, all quantitative
results were produced by executing the verified pipeline rather than by any
generative model, and all edited text was checked for factual accuracy against
our experimental records. We take responsibility for the final content of this
work, including text, claims, and artifacts produced with the aid of
generative AI.

\subsection*{Ethics statement}

This work does not involve human subjects, personally identifiable
information, or crowdsourced annotation. All datasets used are publicly
available research benchmarks (CAD-Recode, Text2CAD, SceneNet, DeepCAD,
Fusion360, T-LESS, MP6D) and are used in accordance with their licenses; our
released dataset consists of synthetic renders of these assets and contains no
private or sensitive content. The primary applications of this
research, reverse-engineering, digitization, and repair of physical parts, are
broadly beneficial; as with any reconstruction technology, it could in
principle be used to copy proprietary parts, but our method offers no
capability beyond what existing measurement and scanning tools already
provide. We foresee no discrimination, fairness, privacy, or security concerns
arising from this work, and we declare no conflicts of interest.

\subsection*{Reproducibility statement}

We have made several efforts to ensure reproducibility. The problem
formulation, reward, and training objectives are fully specified in
Sec.~\ref{sec:method} (Eq.\ref{eq:objective}--\ref{eq:kl_k3}), and the data
generation pipeline, including scene placement, rendering configuration, the
$N{=}36$ view grid, and the visibility/feasibility computation, is described
in Sec.~\ref{sec:data}. The view-selector architecture, SA-max warm-starting
procedure, optimization hyperparameters, the T-LESS/MP6D alignment network,
and the real-capture view-selection protocol are
detailed in Appendix~\ref{sec:app:selector}, and the evaluation protocol,
metrics, and baseline input conventions are given in Sec.~\ref{sec:experiments}
and Appendix~\ref{sec:app:insitu4k}. We have attached the training and evaluation code in the supplementary and will subsequently release the rendered scene-embedded
dataset together with the model checkpoints, so
that all reported results can be regenerated end to end upon acceptance.

\bibliography{iclr2027_conference}
\bibliographystyle{iclr2027_conference}
\newpage
\appendix
\section{Appendix}
In this section we provide additional analysis, results, and framework details that could not be included in the main paper due to space constraints. In the \ref{sec:analysis} subsection we provide a deeper statistical and mechanistic analysis of the learned view selection policy. In \ref{sec:app:qual} we qualitatively illustrate the view inputs and CAD reconstruction outputs for all the baselines and additional datasets compared in the main text. In \ref{sec:app:selector}, we illustrate \textbf{SightCAD} view selector warm-starting training and align-net training for pose estimation for T-LESS and MP6D evaluations, and detail the view-selection protocol applied to the real and canonicalized captures of these benchmarks. In \ref{sec:app:details}, we collect dataset-curation and optimization details. In \ref{sec:app:insitu4k}, we complete the main-paper comparisons on the full scene-embedded test split, evaluating co-trained view policies and external baselines at scale, and report paired significance tests with an occlusion-regime breakdown.

\subsection{View Policy Analysis}
\label{sec:analysis}

In this section we analyze the learned view selection strategy by examining how views change statistically at every budget and with changing part geometry. We also analyze mechanistically why these views perform optimally when input to the VLM CadQuery generator, by first looking at the CadQuery code per-property metrics across views. By ablating the views relevant to the metrics in the activation space of the generator we find these views help infer key properties for the CadQuery code like extrude depth and sketch parameters.

\begin{table}[h]
\centering
\caption{Learned selection policy over the full evaluation set
(4{,}000 test (CAD-Recode and Text2CAD) + 100 DeepCAD + 100 Fusion360, 1365 floor / 977 wall / 1858 corner),
per budget $\Ksel$. \emph{ortho.\ \#} is the number
of views in a selection whose azimuth is cardinal
($0^\circ,90^\circ,180^\circ,270^\circ$ on the 12-azimuth grid), averaged over
objects. \emph{az-gap} is the minimum
pairwise azimuth gap within a selection; \emph{cov.}\ is the fraction of the object's surface area physically visible from the union of the selected views, \emph{rig} is the mean fraction of a selection lying on the modal $\Ksel$-slot rig (the $\Ksel$ globally most frequent view slots of
that policy at that budget). fw-SA is the feature-weighted variant of SA-max
whose greedy coverage objective reweights surface points by local curvature
saliency (Appendix~\ref{sec:app:details}). All columns span the full 4{,}200-object set
for all three policies.}
\label{tab:policy}
\footnotesize
\setlength{\tabcolsep}{3pt}
\resizebox{\linewidth}{!}{%
\begin{tabular}{@{}c cccc cccc cccc@{}}
\toprule
& \multicolumn{4}{c}{Learned} & \multicolumn{4}{c}{SA-max} & \multicolumn{4}{c}{fw-SA} \\
\cmidrule(lr){2-5}\cmidrule(lr){6-9}\cmidrule(lr){10-13}
$\Ksel$ & ortho.\ \# & az-gap ($^\circ$) & cov.\ \% & rig \%
        & ortho.\ \# & az-gap ($^\circ$) & cov.\ \% & rig \%
        & ortho.\ \# & az-gap ($^\circ$) & cov.\ \% & rig \% \\
\midrule
1 & \textbf{0.30} & ---  & 43.5 & 8.7  & 0.16 & ---   & 47.4 & 6.5  & 0.26 & ---   & 45.5 & 5.4  \\
2 & \textbf{0.61} & 52.5 & 54.0 & 16.5 & 0.31 & 135.6 & 75.8 & 9.4  & 0.45 & 131.2 & 73.5 & 9.6  \\
3 & \textbf{1.25} & 15.8 & 60.7 & 21.6 & 0.79 & 55.3  & 81.1 & 16.3 & 0.92 & 55.0  & 80.5 & 17.2 \\
4 & \textbf{2.15} & 11.2 & 66.3 & 39.2 & 1.29 & 31.0  & 82.5 & 25.6 & 1.41 & 29.8  & 82.5 & 26.4 \\
5 & \textbf{2.13} & 12.6 & 72.1 & 43.8 & 1.70 & 21.3  & 82.9 & 34.8 & 1.82 & 20.0  & 83.1 & 36.3 \\
\bottomrule
\end{tabular}%
}
\end{table}

\subsubsection{What do the learned views look like?}

\begin{figure}[h]
\centering
\includegraphics[width=0.87\linewidth]{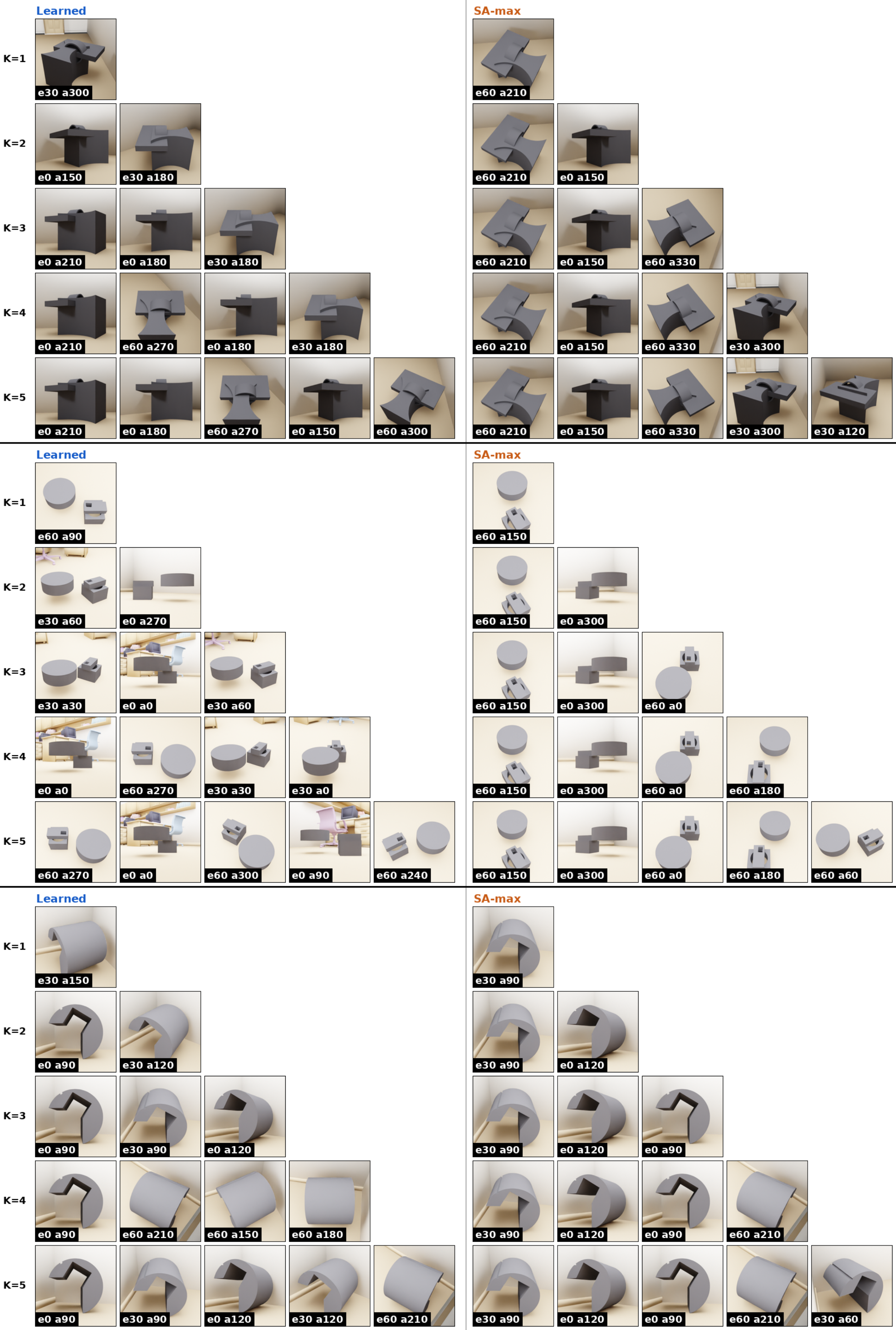}
\caption{Three randomly sampled CAD objects from the view analysis split, showing the learned selector's picks (left)
beside the SA-max chain's first $\Ksel$ picks (right) for every budget $\Ksel\in\{1,\dots,5\}$; each tile is tagged with its (elevation, azimuth) pose. The patterns quantified in the table are directly visible: the learned selector squares views to the cardinal azimuths and stacks elevations at a shared azimuth (small min az-gap), and its per-budget selections are non-nested, whereas
SA-max spreads toward antipodal, oblique, surface-grazing views to maximize coverage.}
\label{fig:policy-qual}
\end{figure}

\paragraph{Learned selection strategy.} Table~\ref{tab:policy} quantifies all
three policies over the full evaluation set (4{,}200 objects), and
Fig.~\ref{fig:policy-qual} shows representative selections side by side at
every budget. The learned selector allocates its budget like an engineering
draftsman, and its strategy changes qualitatively with $\Ksel$. Below the
budget needed for a full drawing, selection is \emph{feature-centric}: the
whole budget is spent on the single region that fixes the most program
parameters. At $\Ksel{=}1$ it takes a single mid-elevation
(canonical) recognition view~\citep{palmer1981cannonical} that maximizes the gross-shape visibility of the part, similar (but not identical) to the view chosen by SA-max. At $\Ksel{=}2$ it does not add a complementary face but a second view close in azimuth: a narrow stereo \emph{wedge} straddling the same feature to pin its depth, plainly visible in
the $\Ksel{=}2$ rows of Fig.~\ref{fig:policy-qual} and in the min az-gap column ($52.5^\circ$ against SA-max's near-antipodal $135.6^\circ$). Only from $\Ksel{=}3$, with the feature secured, does an axis-aligned reading appear: an orthographic front profile, a top-down plan, and a three-quarter
anchor similar to the projection triad of a mechanical drawing. The ortho.\ \#\ column traces this transition: the learned cardinal
supply grows from $0.30$ picks per selection at $\Ksel{=}1$ to $2.15$ at $\Ksel{=}4$, a majority of the budget and $1.6$--$2\times$ SA-max's supply for $\Ksel\leq4$ ($1.29$ at $\Ksel{=}4$), narrowing to $1.25\times$ at $\Ksel{=}5$: scoring area alone, SA-max prefers the oblique azimuths that graze the most surface. Simultaneously the min az-gap falls to ${\sim}11^\circ$ as the selector stacks elevations at a shared azimuth to read a single face at several heights (the same-azimuth tile columns in Fig.~\ref{fig:policy-qual}). At $\Ksel{=}5$ the cardinal supply plateaus ($2.13$ from $2.15$): the fifth learned view departs from the cardinal angles. This view legibility by the learned policy is bought at the expense of visible area coverage over the scene-embedded CAD part: in the cov.\ columns, learned subsets range from $43.5$ to $72.1\%$ of the object's surface across $\Ksel\in\{1,\dots,5\}$ versus $47.4$ to $82.9\%$ for SA-max. SA-max essentially saturates by $\Ksel{=}3$ ($81.1\%$ to ${\approx}83\%$ ceiling) as the scene-embedded setting fundamentally hides part surface from all possible view angles. Both policies drift toward different templates as the budget grows (rig $6.5$ to $34.8\%$ SA-max, $8.7$ to $43.8\%$ learned in Table~\ref{tab:policy}): SA-max's view rig is entirely geometry-centric whereas the learned modal rig template is partly drawing-rig-centric (with the cardinal views at $\Ksel{\geq}3$) and partly geometry-centric. The fw-SA columns show that curvature reweighting only nudges these statistics within the coverage family: fw-SA changes the selected set for roughly two thirds of objects, yet its cardinal supply, az-gap, coverage, and rig share all track SA-max closely (e.g.\ ortho.\ $1.41$ vs.\ $1.29$ at $\Ksel{=}4$, coverage within $2.3$ pp everywhere) and stay far from the learned profile, mirroring the IoU tie of Table~\ref{tab:schemes-main}. We examine this geometry dependence in the next paragraph.

\begin{figure}[h]
\centering
\includegraphics[width=0.875\linewidth]{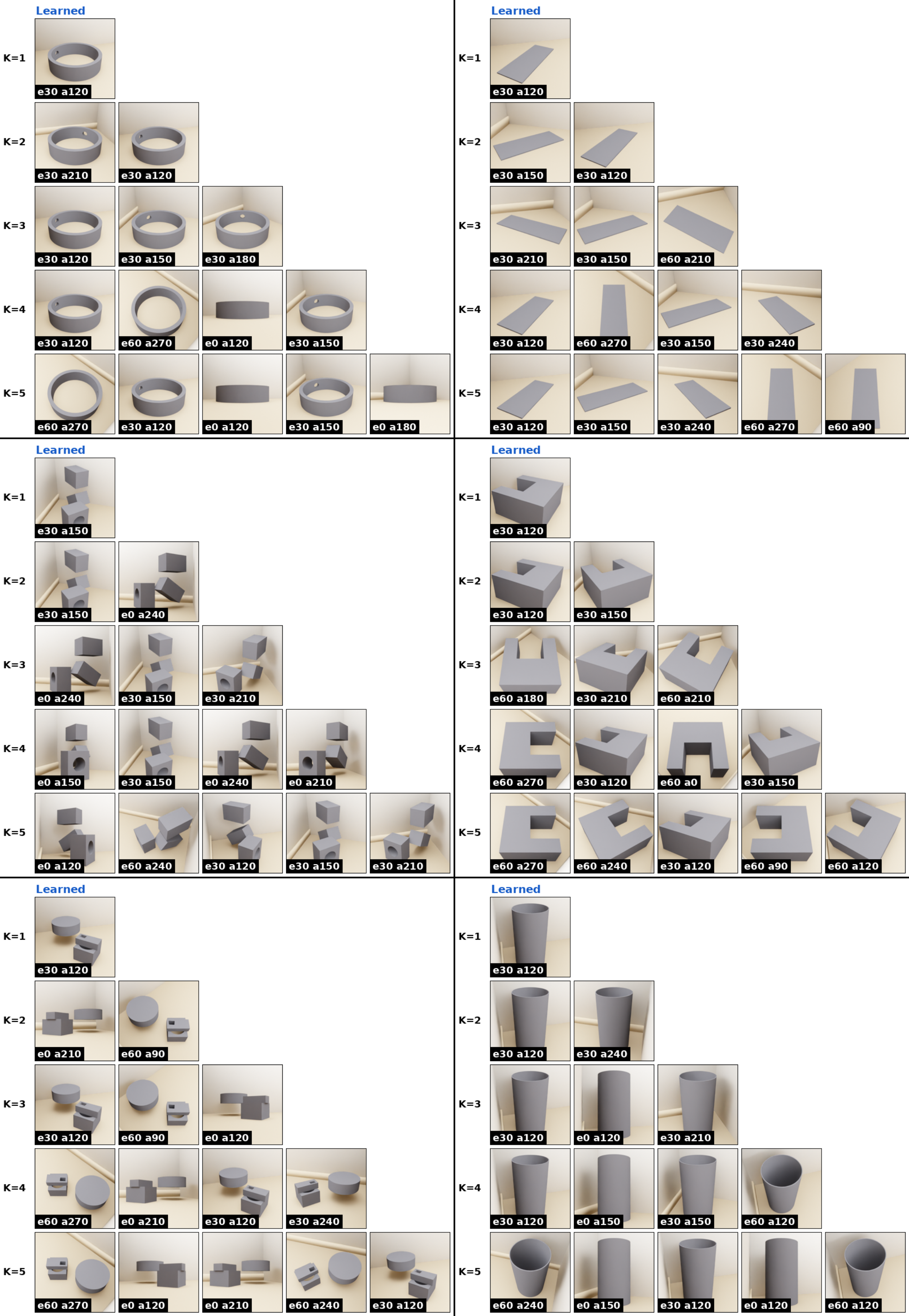}
\caption{Controlled same-scene study of geometry-dependent selection. Six
parts spanning the geometry taxonomy: a ring with a lateral through-hole and a flat plate (top row), a block stack pierced by a circular bore and a U-bracket (middle row), a cylinder--block assembly and a hollow cylinder (bottom row); are placed in the \emph{same} room at the \emph{same} corner
anchor with an \emph{identical} shared yaw of $45^\circ$. Each panel shows the frozen per-$\Ksel$ selectors' picks
for budgets $\Ksel\in\{1,\dots,5\}$; each tile is tagged with its (elevation,
azimuth) pose. With scene, anchor, lighting, and feasibility held fixed, all
differences between panels are attributable to part geometry alone.}
\label{fig:policy-fixedscene}
\end{figure}

\paragraph{Dependence on part geometry.}
A natural concern is that the learned policy might collapse to a fixed
per-$\Ksel$ template. Our finding here is contrary: while selection tilts toward a shared
canonical rig as the budget grows (rig share $8.7\%$ at $\Ksel{=}1$ rising to
$43.8\%$ at $\Ksel{=}5$, Table~\ref{tab:policy}), object-specific picks
remain the majority at every budget, and
Fig.~\ref{fig:policy-fixedscene} shows that these picks are dictated by
part geometry alone. There, six parts of contrasting geometry are placed in
the \emph{same} room, at the \emph{same} corner anchor, with the same
$45^\circ$ yaw. The selections diverge sharply beyond $\Ksel{=}1$ (which largely converges to the \texttt{e30\_a120} view). The ring
brackets its lateral through-hole with a narrow \texttt{e30} wedge before adding a
plan straight down the bore; the flat plate is read only from mid and high
elevations, never from a level view that would foreshorten it to a sliver. The bore-pierced block stack gets level front views
that project its bore as a true circle while the U-bracket's budget goes almost entirely to \texttt{e60} plans looking down into the channel. The two-piece
assembly splits its picks between plans over the cylinder and level views of the slotted block; and the hollow cylinder receives near-antipodal level profiles for height and diameter plus late rim views that reveal it is hollow. The same selector, in
the same scene, thus reallocates its budget from feature bracketing (ring,
bore, channel) to profile reading (plate, cylinder) to member splitting
(assembly) purely as a function of part geometry, and the same pattern recurs across the test dataset.

\subsubsection{Why does the VLM prefer these views?}
\label{sec:mechanism}

The preceding paragraphs described \emph{what} the learned policy selects; in this section we explore \emph{why} those selections help. All measurements here use the policy-neutral generator of Table~\ref{tab:schemes-main}b, so every effect is a property of the views.

\begin{figure}[h]
\centering
\includegraphics[width=\linewidth]{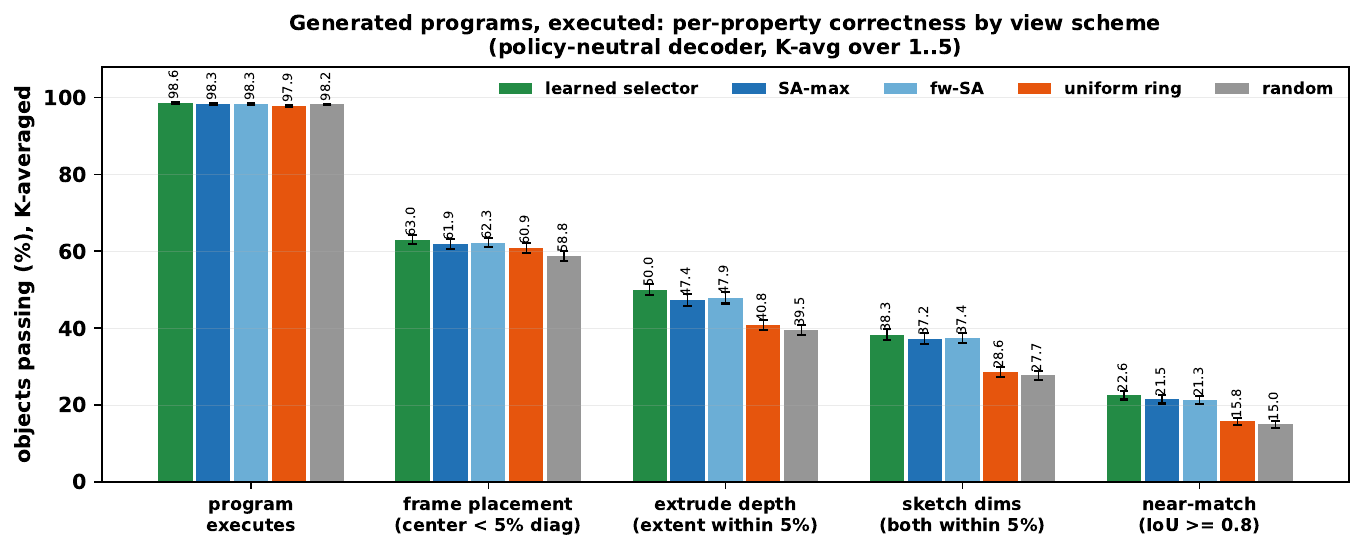}
\caption{Executed per-property correctness under five view-selection policies
decoded by the same policy-neutral generator, averaged over $\Ksel\in\{1,\dots,5\}$. Program validity is view-policy independent; the graded properties degrade in the
ordering of Table~\ref{tab:schemes-main}b; extrude depth is the major property on
which learned selection separates from SA-max.}
\label{fig:exec-props}
\end{figure}

\textbf{Where the views act.} We first decompose the end-to-end gap by
scoring the \emph{executed} program property-by-property
(Fig.~\ref{fig:exec-props}). Program structure is view-independent: every
policy, unrestricted random included, emits executable code
$97.9$--$98.6\%$ of the time. We find that different view policies affect numeric parameters generated by the policy-neutral VLM differently, of which \emph{extrude depth} and \emph{sketch parameters} show the largest differences between the view policies. 

\begin{figure}[h]
\centering
\begin{subfigure}[t]{0.5\linewidth}
\centering
\includegraphics[width=\linewidth]{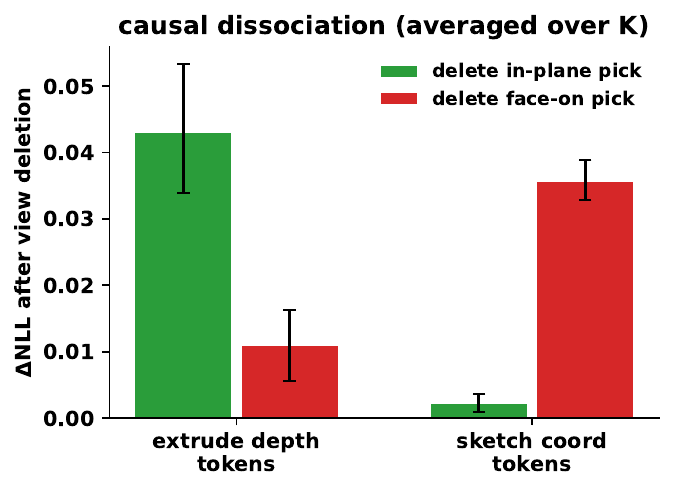}
\caption{Causal dissociation.}
\label{fig:kavg-dissoc}
\end{subfigure}\hfill
\begin{subfigure}[t]{0.5\linewidth}
\centering
\includegraphics[width=\linewidth]{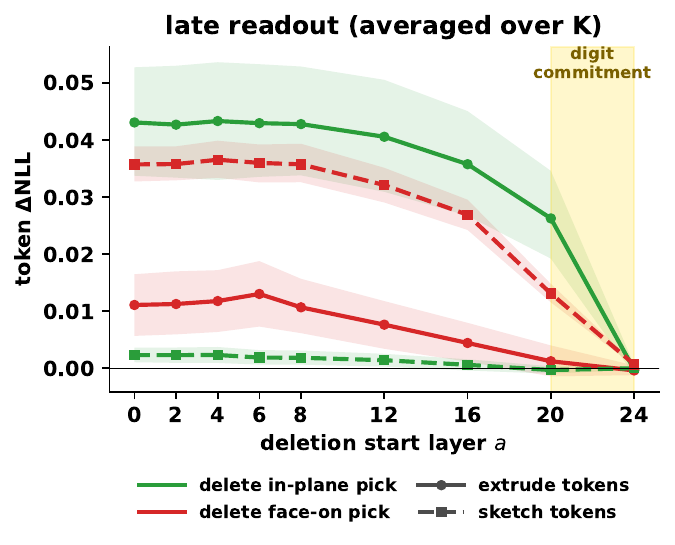}
\caption{Late readout.}
\label{fig:kavg-readout-combined}
\end{subfigure}

\caption{\textbf{The view-class mechanism, averaged over budgets
$\Ksel$} (learned-selector picks, token-column ablation with
other-views-mean replacement).
(a)~Deleting the in-plane pick damages extrude-depth tokens while deleting
the face-on pick damages sketch-coordinate tokens.
(b)~Late-readout analysis: extrude-token damage from deleting the in-plane
pick remains approximately flat through L16 and collapses to zero by L24,
while sketch-token damage from deleting the face-on pick follows the same
late collapse around the logit-lens digit-commitment window. Depth and
sketch parameters are thus read late, from disjoint view classes, at every
budget $\Ksel$.}
\label{fig:mechanism-summary}
\end{figure}

\textbf{Mechanism.} 
We divide the views of the CAD geometry into two categories to find the relationship of views with depth and sketch parameters: \emph{face-on} and \emph{in-plane}, where the camera view angles are $\leq15^\circ$ and $\geq75^\circ$ to the part's sketch normal, respectively. To find this causal relationship between views and key numerical parameters (extrude depth and sketch parameters) we do the following: (i) We first find the view's visual tokens. As the $\Ksel$ views are concatenated into one wide image in the visual encoder of our VLM (Qwen2-VL 2B), a view owns contiguous column ranges of the token grid. (ii) We then ablate the view tokens pertaining to both face-on and in-plane views in our horizontal view input strip at every layer in the decoder's residual stream, by replacing them with the mean of other views' tokens (so that the hidden states stay in-distribution) at that layer. (iii) We then teacher-force the ground truth CadQuery program and measure the $\Delta\mathrm{NLL}$ restricted to the specific token categories, namely extrude-depth digits and sketch-coordinate digits. We find that, averaged across $\Ksel\in\{1,\dots,5\}$, there is a causal dissociation between in-plane views and extrude-depth tokens, and between face-on views and sketch parameters. This is visible in Fig.~\ref{fig:mechanism-summary}a: the extrude-depth NLL and the sketch-parameter NLL degrade most sharply when the in-plane and face-on tokens, respectively, are removed at the input layer of the decoder. Figures~\ref{fig:mechanism-summary}b--c show the timing: the $\Delta\mathrm{NLL}$ from ablating the in-plane (resp.\ face-on) tokens stays flat through layer 16 and collapses to zero by layer 24, the logit-lens window in which the digit tokens are committed, indicating a late readout of both properties.

\begin{figure}[h!]
    \centering
    \includegraphics[width=0.7\linewidth]{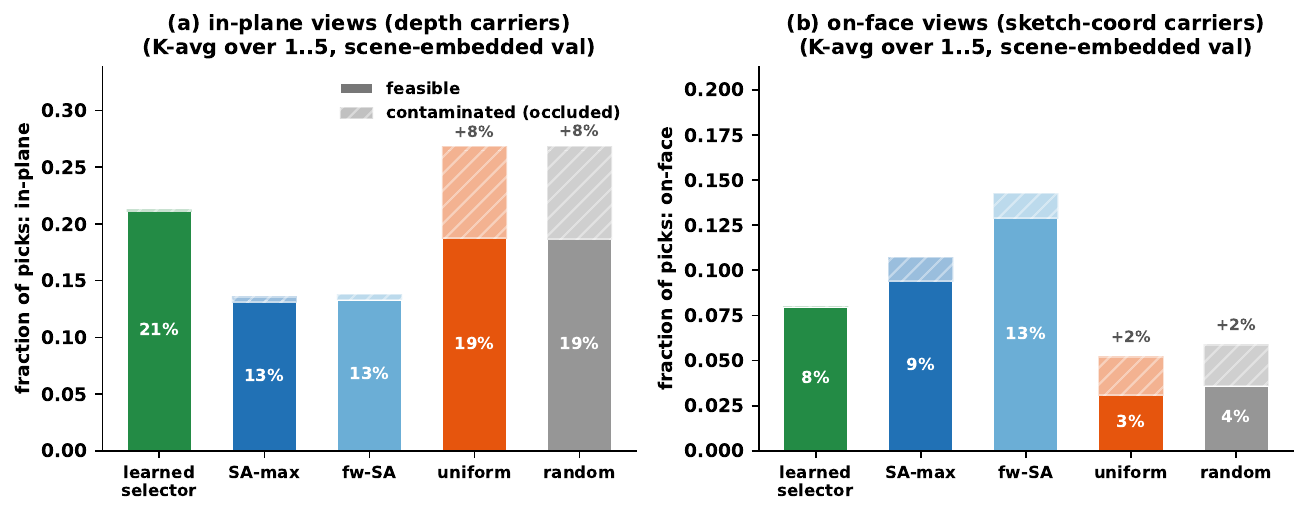}
    \caption{In-plane and face-on view supply for all view selection policies considered in this work. Occluded (contaminated) views, together with the frequency of in-plane and face-on views, correlate with the loss in accuracy on the extrude depth and sketch parameters, respectively, in Fig.~\ref{fig:exec-props}.}
    \label{fig:supply_bars}
\end{figure}

\textbf{Conclusion}
Finally, the supply of in-plane and face-on views under each view selection policy (accounting also for the occluded views drawn by the random and uniform policies), shown in Fig.~\ref{fig:supply_bars}, correlates strongly with the drop in accuracy on the extrude depth and sketch parameters (Fig.~\ref{fig:exec-props}), consistent with the mechanism observed above. This confirms to an extent that learned policy wins by supplying the depth and sketch-parameter carriers that other view selection policies under-supply.

\clearpage
\subsection{Qualitative Results}
\label{sec:app:qual}
\begin{figure}[h!]
\centering
\includegraphics[width=\linewidth]{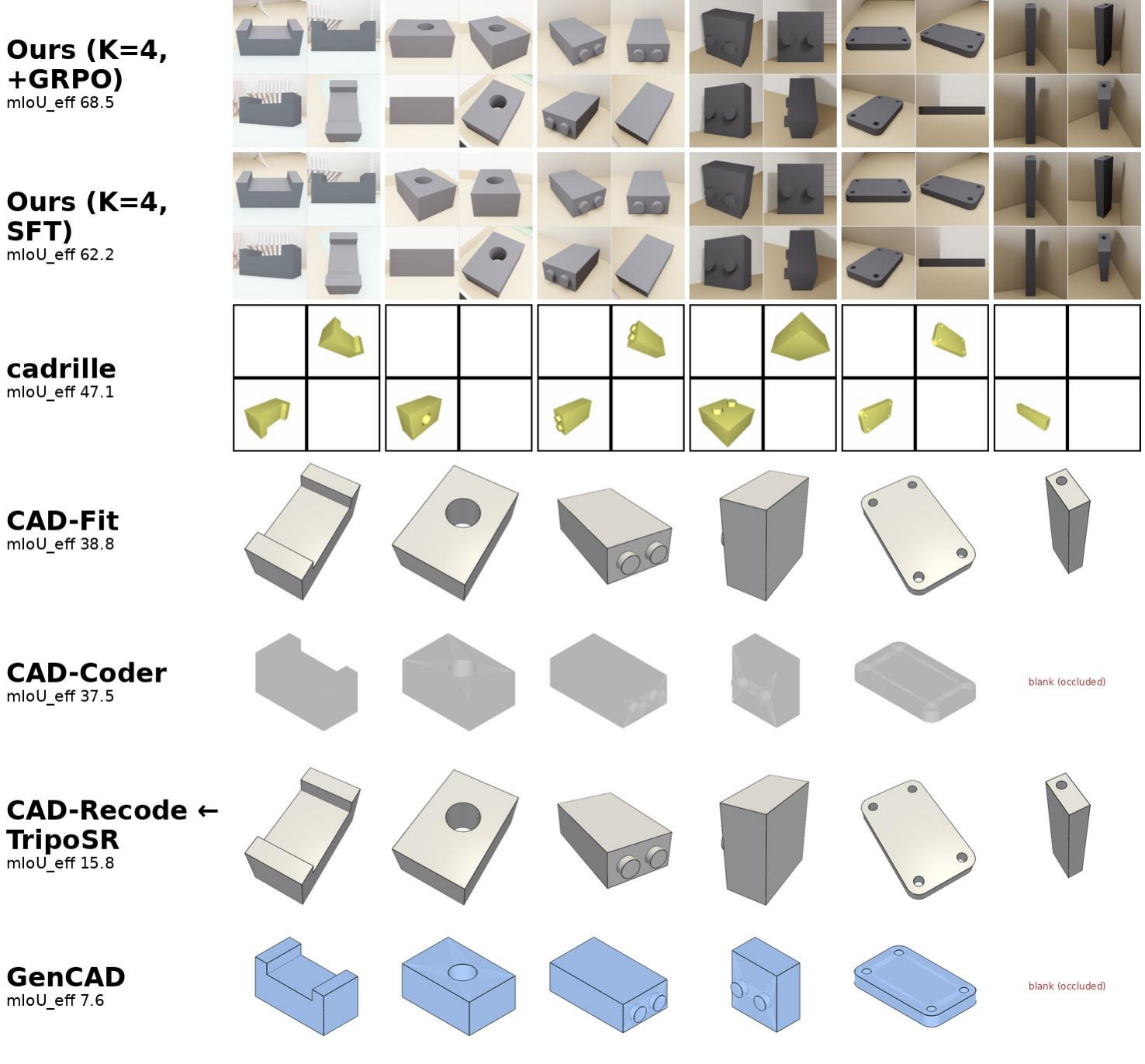}
\caption{\textbf{DeepCAD-test model inputs.} For eight held-out
objects, the exact visual input consumed by each method: \emph{Ours} receives
the four scene-embedded views chosen by the learned selector (2$\times$2 montage);
\emph{cadrille} its native yellow four-quadrant view grid with scene-occluded quadrants
whited out; \emph{CAD-Fit} and \emph{CAD-Recode$\leftarrow$TripoSR} their
native white-background single-view studio render; \emph{CAD-Coder} and
\emph{GenCAD} their native isometric / edge-lattice render, blanked where the
scene occludes the native camera. Rows are ordered by mIoU\textsubscript{eff}.}
\label{fig:qual-inputs-deepcad-v2}
\end{figure}

\begin{figure}[h!]
\centering
\includegraphics[width=\linewidth]{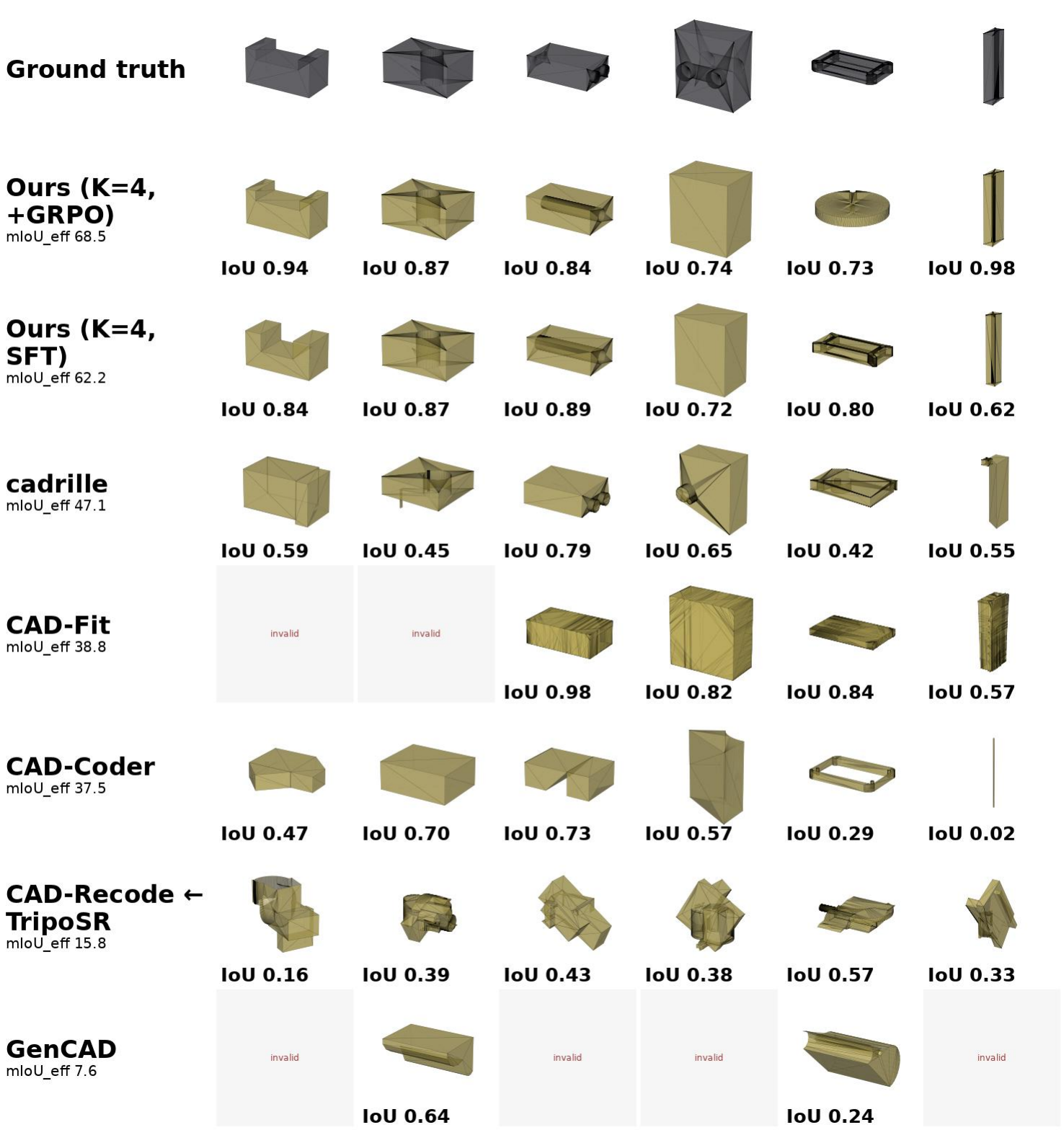}
\caption{\textbf{DeepCAD-test reconstructions.} Executed CAD for
the objects of Fig.~\ref{fig:qual-inputs-deepcad-v2}: ground truth (top) and
each method's solid, rendered semi-transparent so internal features are
visible and aligned by its per-object best-fit proper rotation (the same
rotation scores the printed IoU), so shape quality is shown independent of
coordinate-frame convention.}
\label{fig:qual-preds-deepcad-v2}
\end{figure}

\begin{figure}[h!]
\centering
\includegraphics[width=\linewidth]{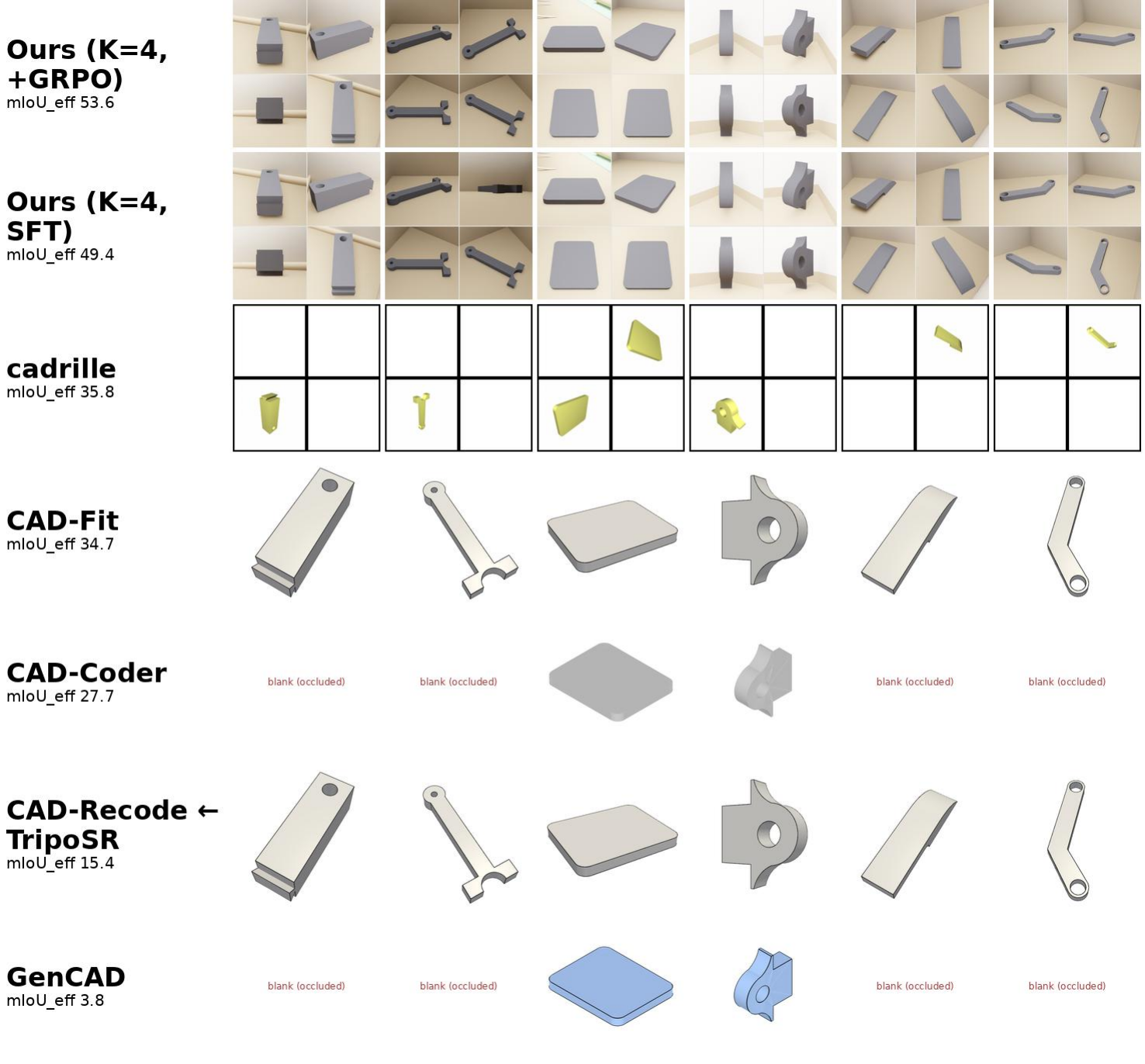}
\caption{\textbf{Fusion360-test model inputs.} For eight held-out
objects, the exact visual input consumed by each method: \emph{Ours} receives
the four scene-embedded views chosen by the learned selector (2$\times$2 montage);
\emph{cadrille} its native yellow four-quadrant view grid with scene-occluded quadrants
whited out; \emph{CAD-Fit} and \emph{CAD-Recode$\leftarrow$TripoSR} their
native white-background single-view studio render; \emph{CAD-Coder} and
\emph{GenCAD} their native isometric / edge-lattice render, blanked where the
scene occludes the native camera. Rows are ordered by mIoU\textsubscript{eff}.}
\label{fig:qual-inputs-fusion360-v2}
\end{figure}

\begin{figure}[h!]
\centering
\includegraphics[width=\linewidth]{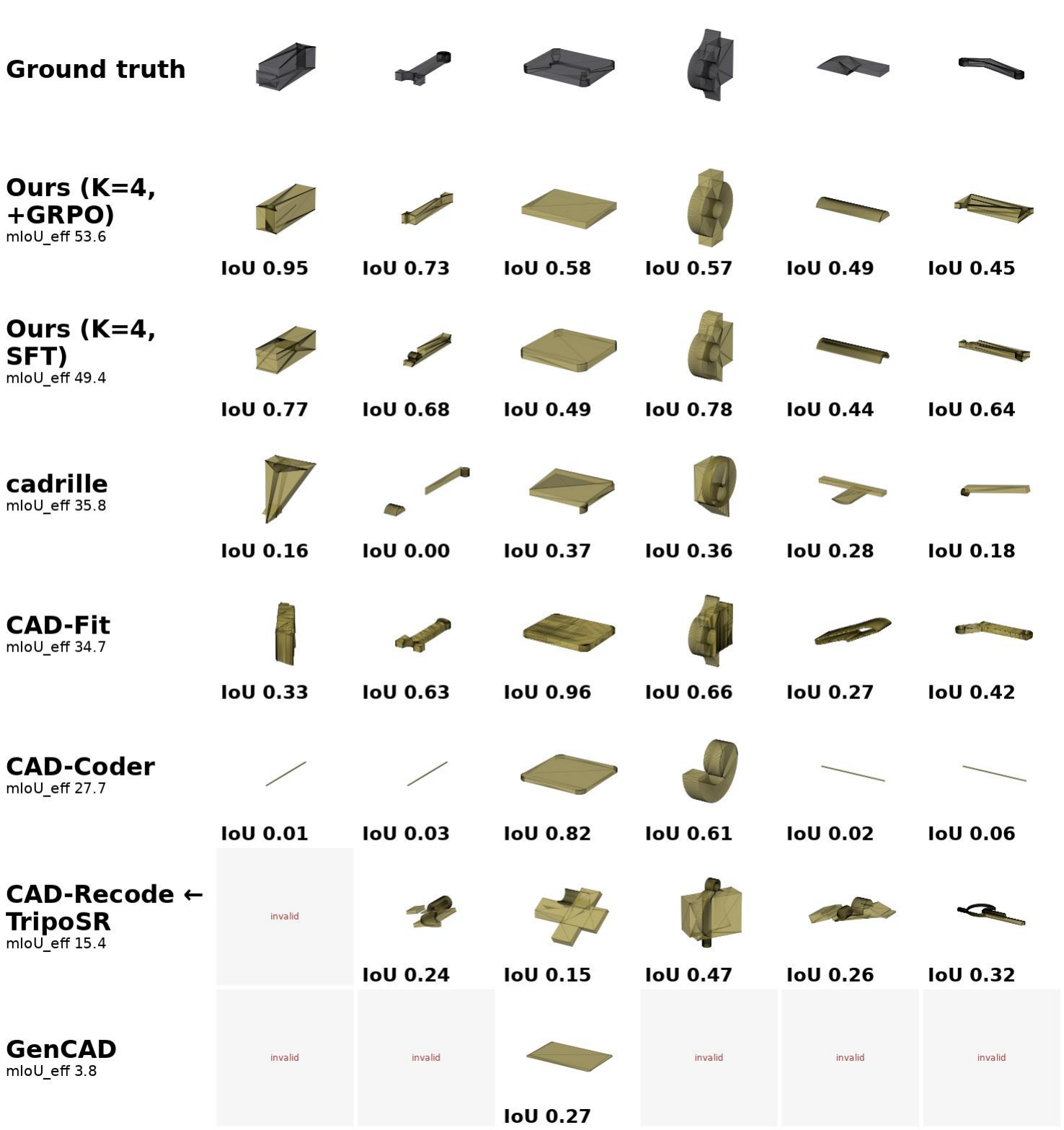}
\caption{\textbf{Fusion360-test reconstructions.} Ground truth and
per-method executed solids for the objects of
Fig.~\ref{fig:qual-inputs-fusion360-v2}, shown translucent and aligned by each
object's best-fit rotation with the corresponding IoU.}
\label{fig:qual-preds-fusion360-v2}
\end{figure}

\begin{figure}[h!]
\centering
\includegraphics[width=\linewidth]{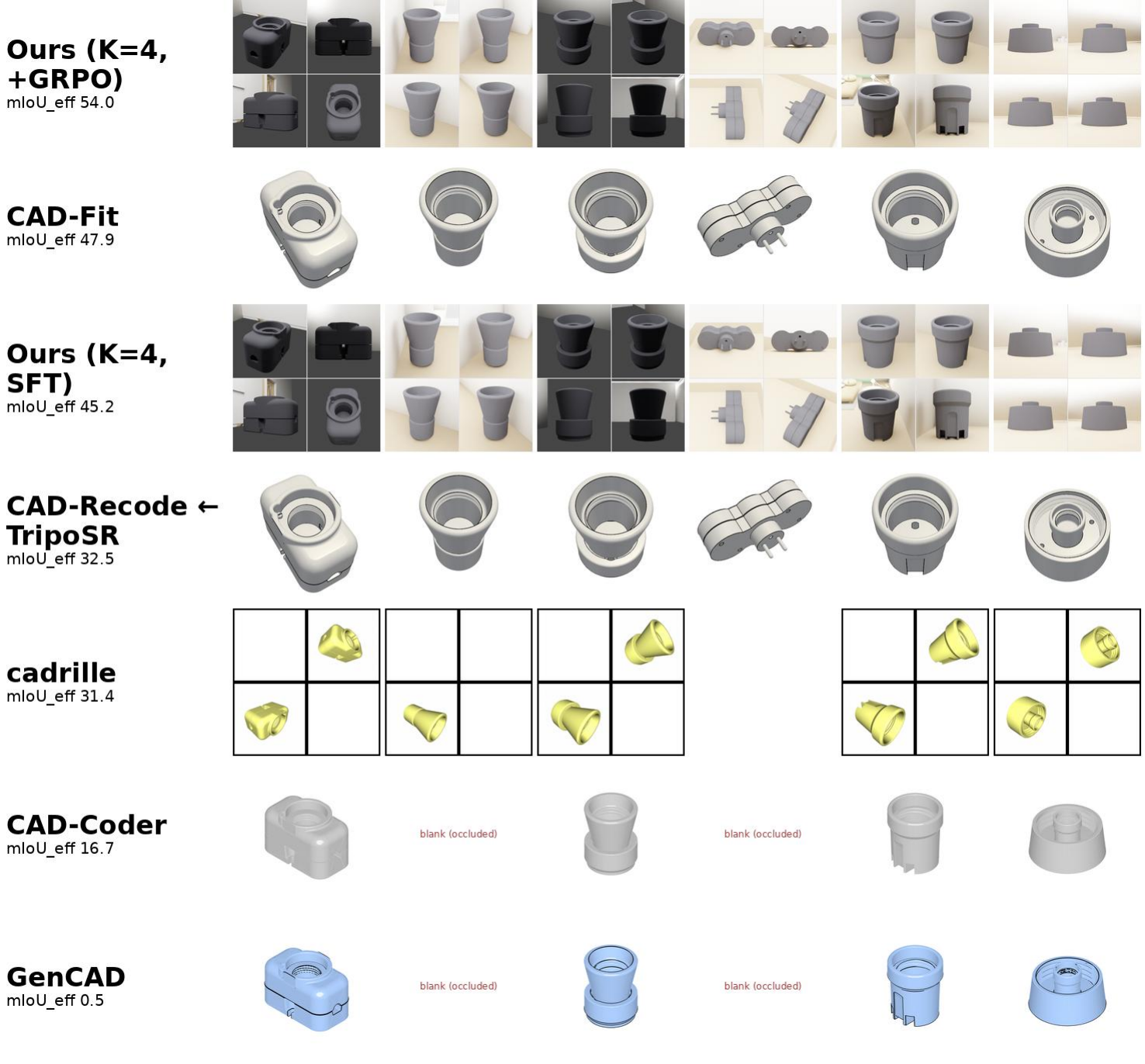}
\caption{\textbf{T-LESS synth model inputs.} For eight held-out
objects, the exact visual input consumed by each method: \emph{Ours} receives
the four scene-embedded views chosen by the learned selector (2$\times$2 montage);
\emph{cadrille} its native yellow four-quadrant view grid with scene-occluded quadrants
whited out; \emph{CAD-Fit} and \emph{CAD-Recode$\leftarrow$TripoSR} their
native white-background single-view studio render; \emph{CAD-Coder} and
\emph{GenCAD} their native isometric / edge-lattice render, blanked where the
scene occludes the native camera. Rows are ordered by mIoU\textsubscript{eff}.}
\label{fig:qual-inputs-tlesssynth-v2}
\end{figure}

\begin{figure}[h!]
\centering
\includegraphics[width=\linewidth]{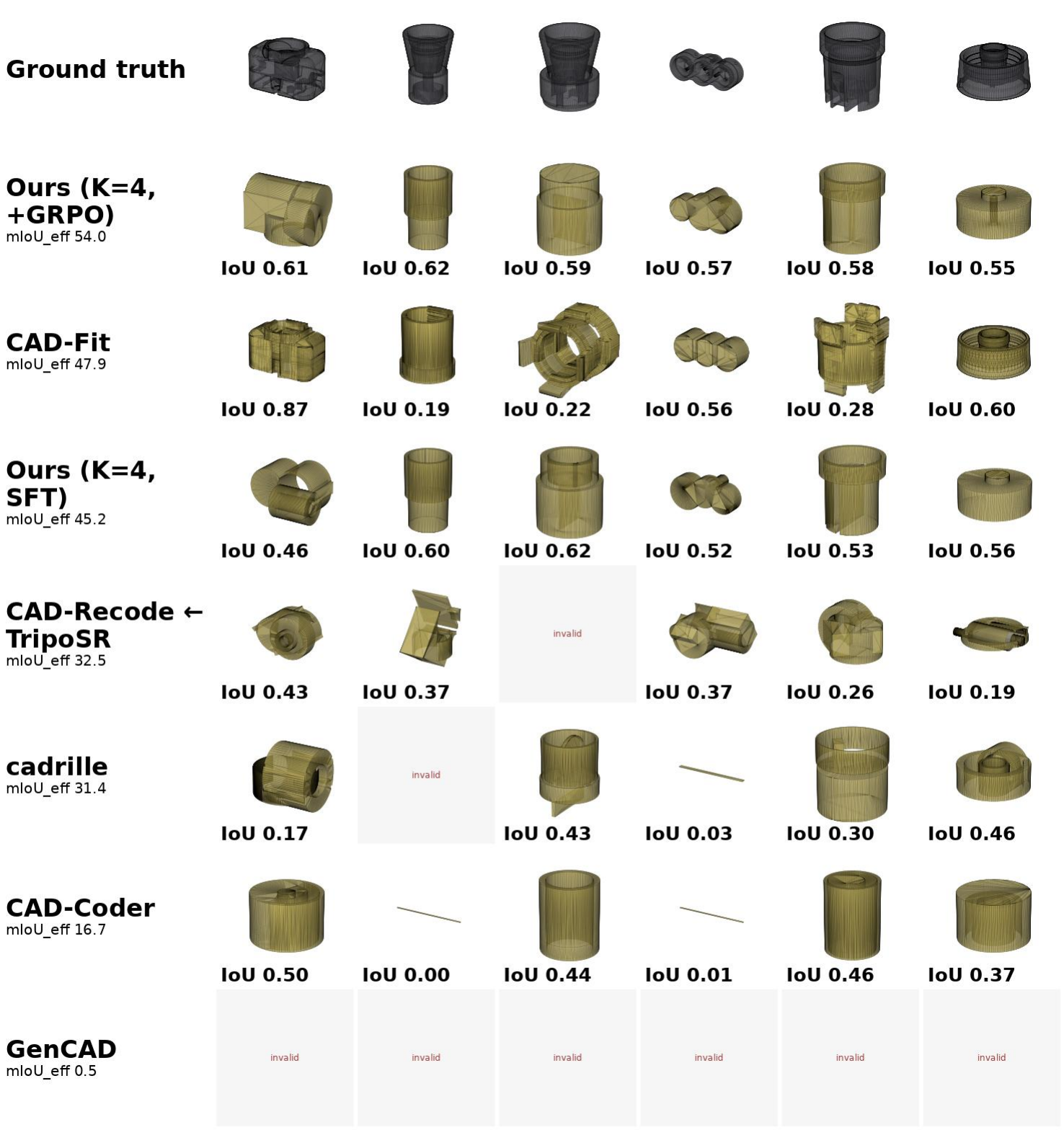}
\caption{\textbf{T-LESS synth reconstructions.} Ground truth and
per-method executed solids for the objects of
Fig.~\ref{fig:qual-inputs-tlesssynth-v2}, translucent and aligned by each
object's best-fit rotation with the corresponding IoU.}
\label{fig:qual-preds-tlesssynth-v2}
\end{figure}

\begin{figure}[h!]
\centering
\includegraphics[width=\linewidth]{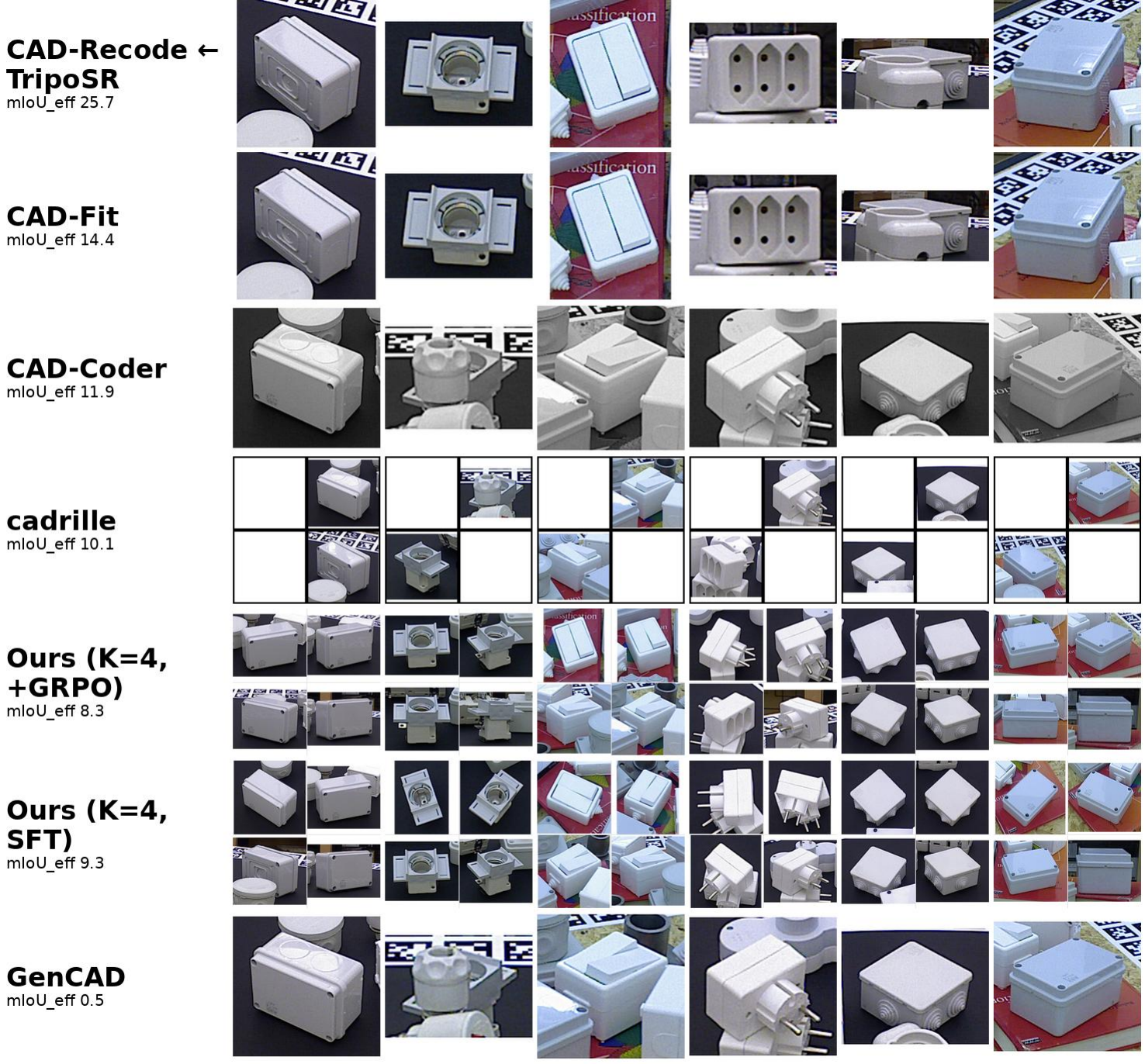}
\caption{\textbf{T-LESS real model inputs.} The real
photographs fed to each method for eight T-LESS captures, without any
test-time domain match (conventions as in
Fig.~\ref{fig:qual-inputs-deepcad-v2}); ours consumes the real scene-embedded views
directly.}
\label{fig:qual-inputs-tlessraw-v2}
\end{figure}

\begin{figure}[h!]
\centering
\includegraphics[width=\linewidth]{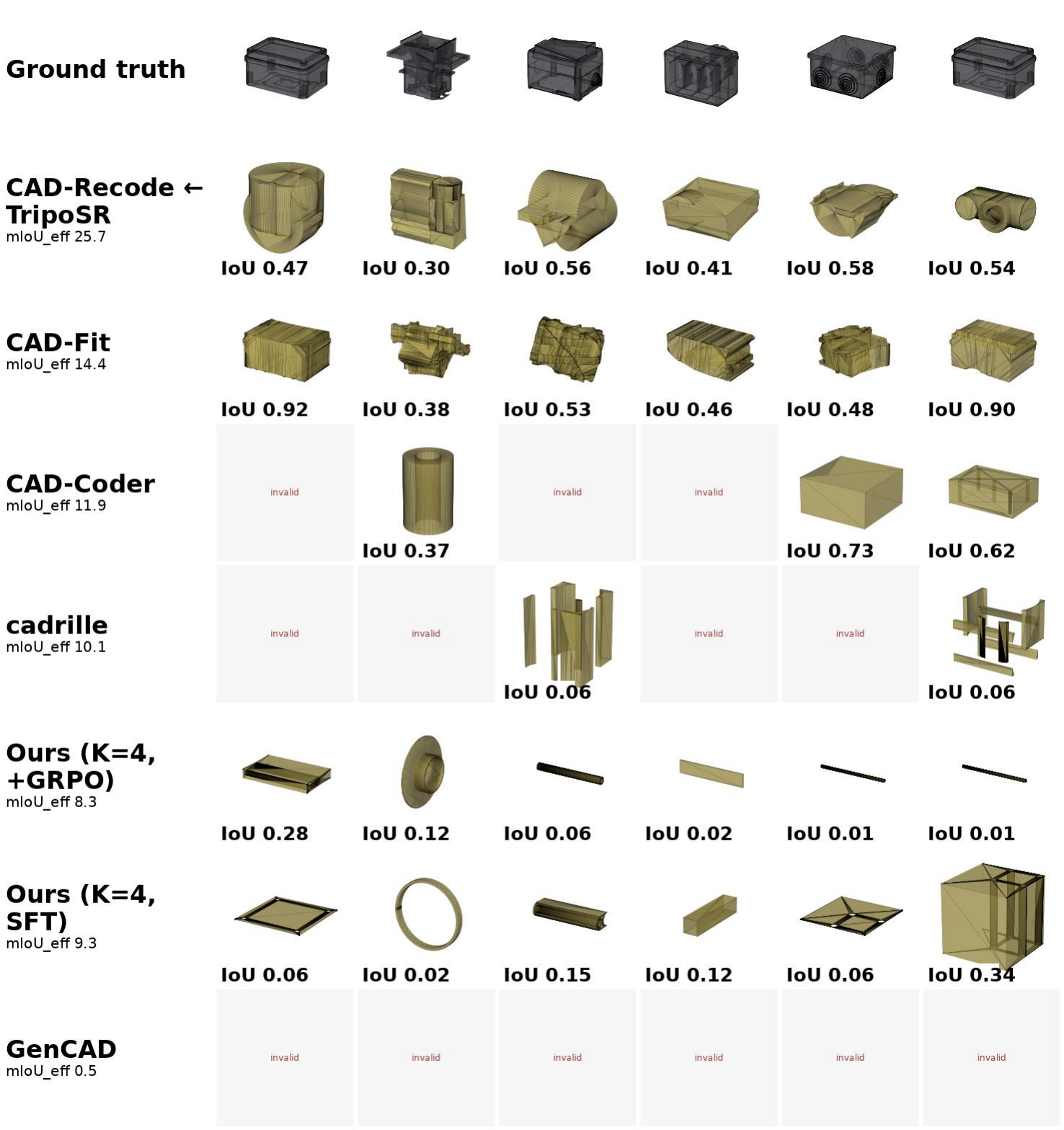}
\caption{\textbf{T-LESS real reconstructions.} Ground truth
and per-method executed solids for the objects of
Fig.~\ref{fig:qual-inputs-tlessraw-v2}, translucent and aligned by each
object's best-fit rotation with the corresponding IoU. This is the one regime
where our framework trails the lift baselines, reflecting the raw-photograph
domain gap discussed in the main text.}
\label{fig:qual-preds-tlessraw-v2}
\end{figure}

\begin{figure}[h!]
\centering
\includegraphics[width=\linewidth]{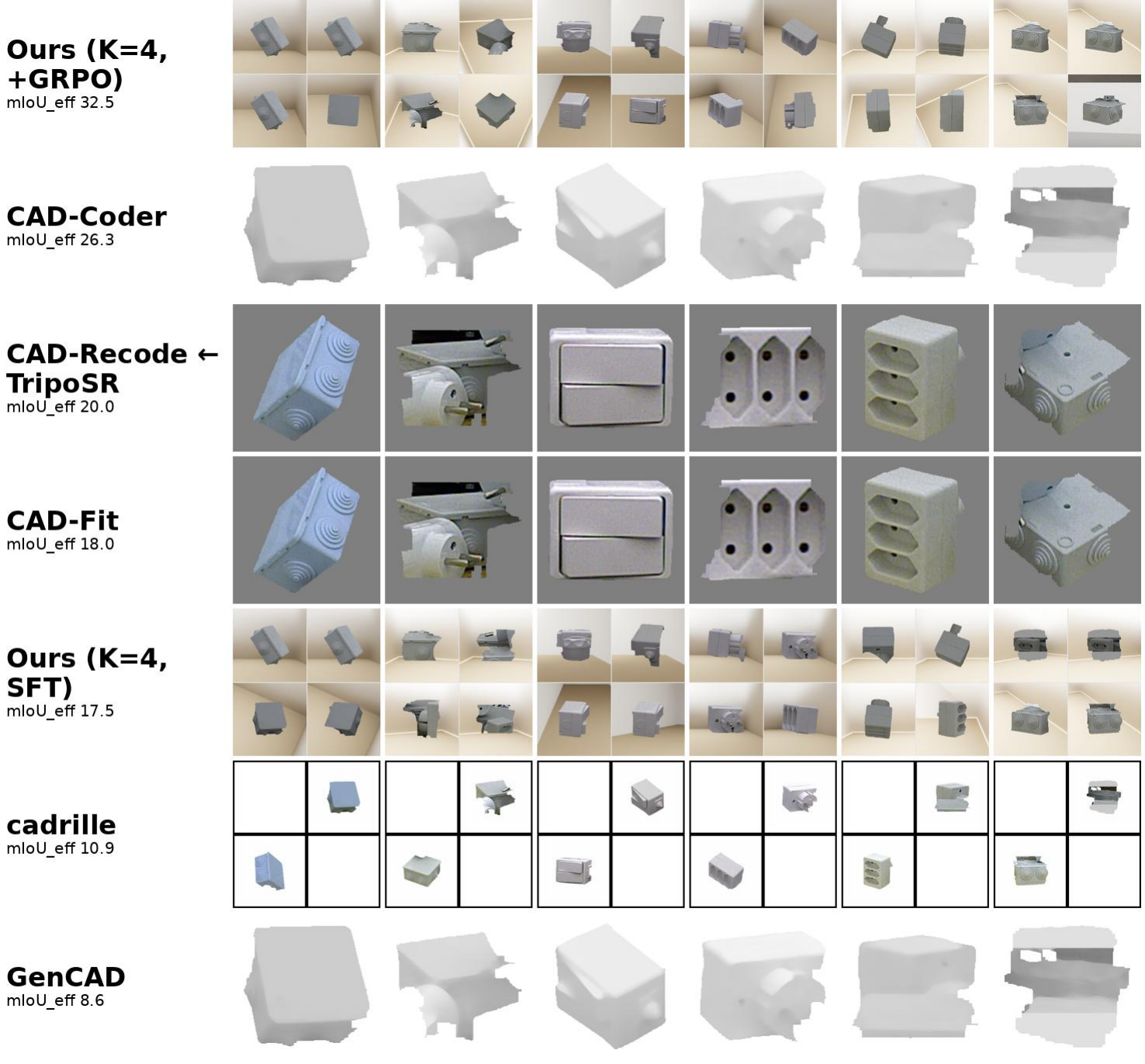}
\caption{\textbf{T-LESS real (canon) model inputs.} The same real
captures as Fig.~\ref{fig:qual-inputs-tlessraw-v2} after the model-agnostic,
test-time-only domain match, composited onto each method's native plate
(conventions as in Fig.~\ref{fig:qual-inputs-deepcad-v2}).}
\label{fig:qual-inputs-tlesscanon-v2}
\end{figure}

\begin{figure}[h!]
\centering
\includegraphics[width=\linewidth]{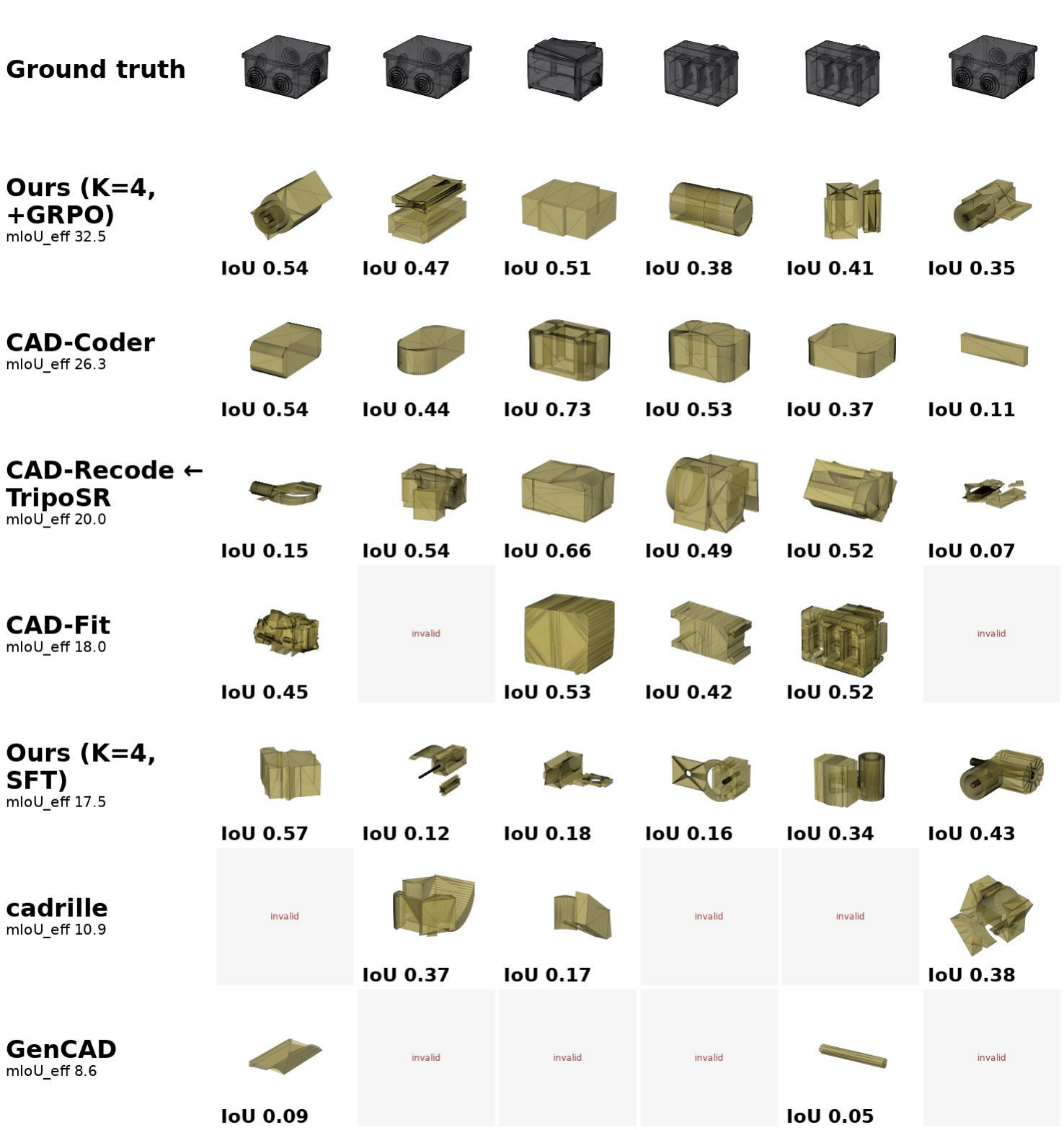}
\caption{\textbf{T-LESS real (canon) reconstructions.} Ground
truth and per-method executed solids for the objects of
Fig.~\ref{fig:qual-inputs-tlesscanon-v2}, translucent and aligned by each
object's best-fit rotation with the corresponding IoU.}
\label{fig:qual-preds-tlesscanon-v2}
\end{figure}

\begin{figure}[h!]
\centering
\includegraphics[width=\linewidth]{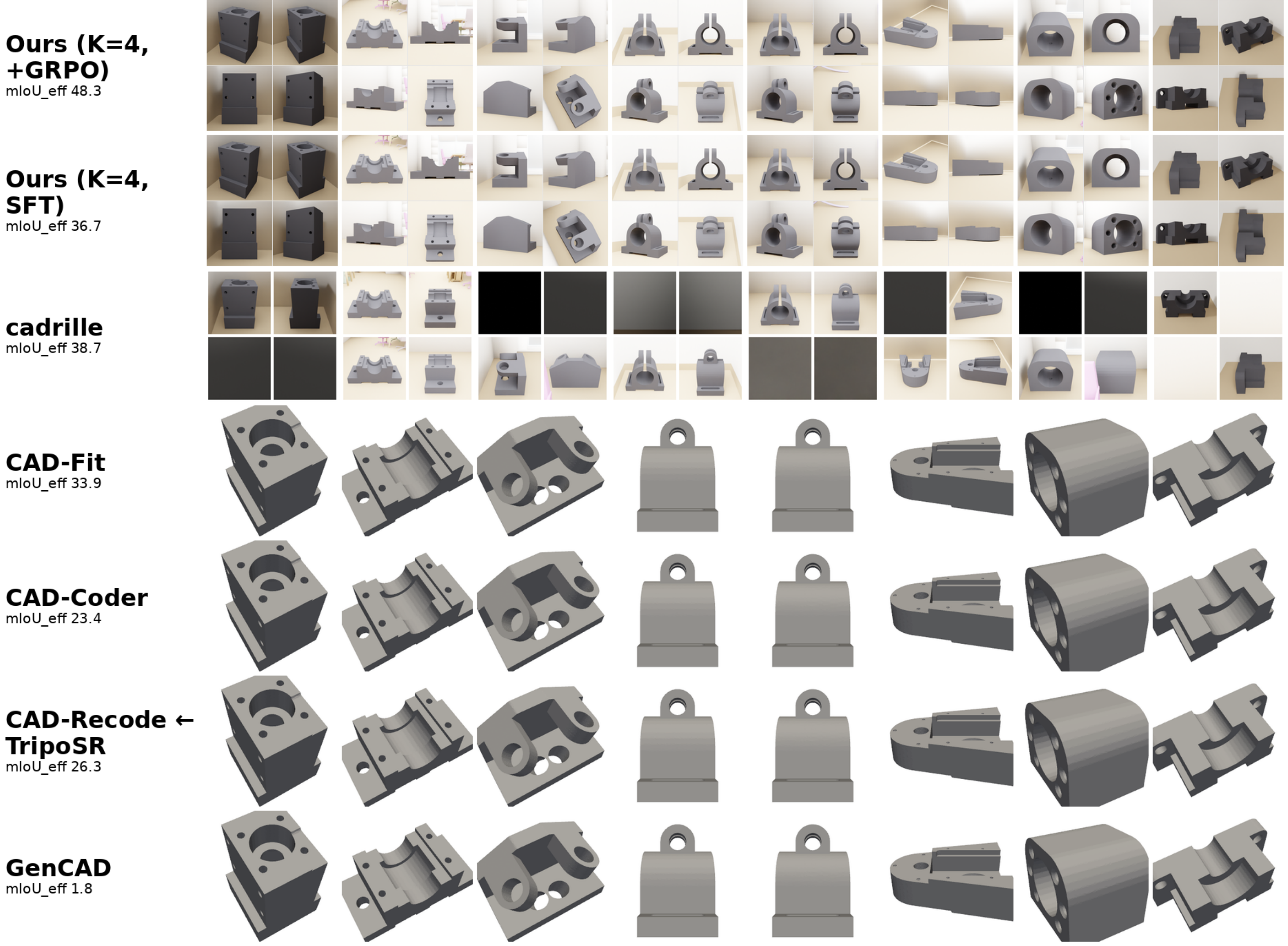}
\caption{\textbf{MP6D synth model inputs.} For eight held-out
objects, the exact visual input consumed by each method: \emph{Ours} receives
the four scene-embedded views chosen by the learned selector (2$\times$2 montage);
\emph{cadrille} its native yellow four-quadrant view grid with scene-occluded quadrants
whited out; \emph{CAD-Fit} and \emph{CAD-Recode$\leftarrow$TripoSR} their
native white-background single-view studio render; \emph{CAD-Coder} and
\emph{GenCAD} their native isometric / edge-lattice render, blanked where the
scene occludes the native camera. Rows are ordered by mIoU\textsubscript{eff}.}
\label{fig:qual-inputs-mp6dsynth-v2}
\end{figure}

\begin{figure}[h!]
\centering
\includegraphics[width=\linewidth]{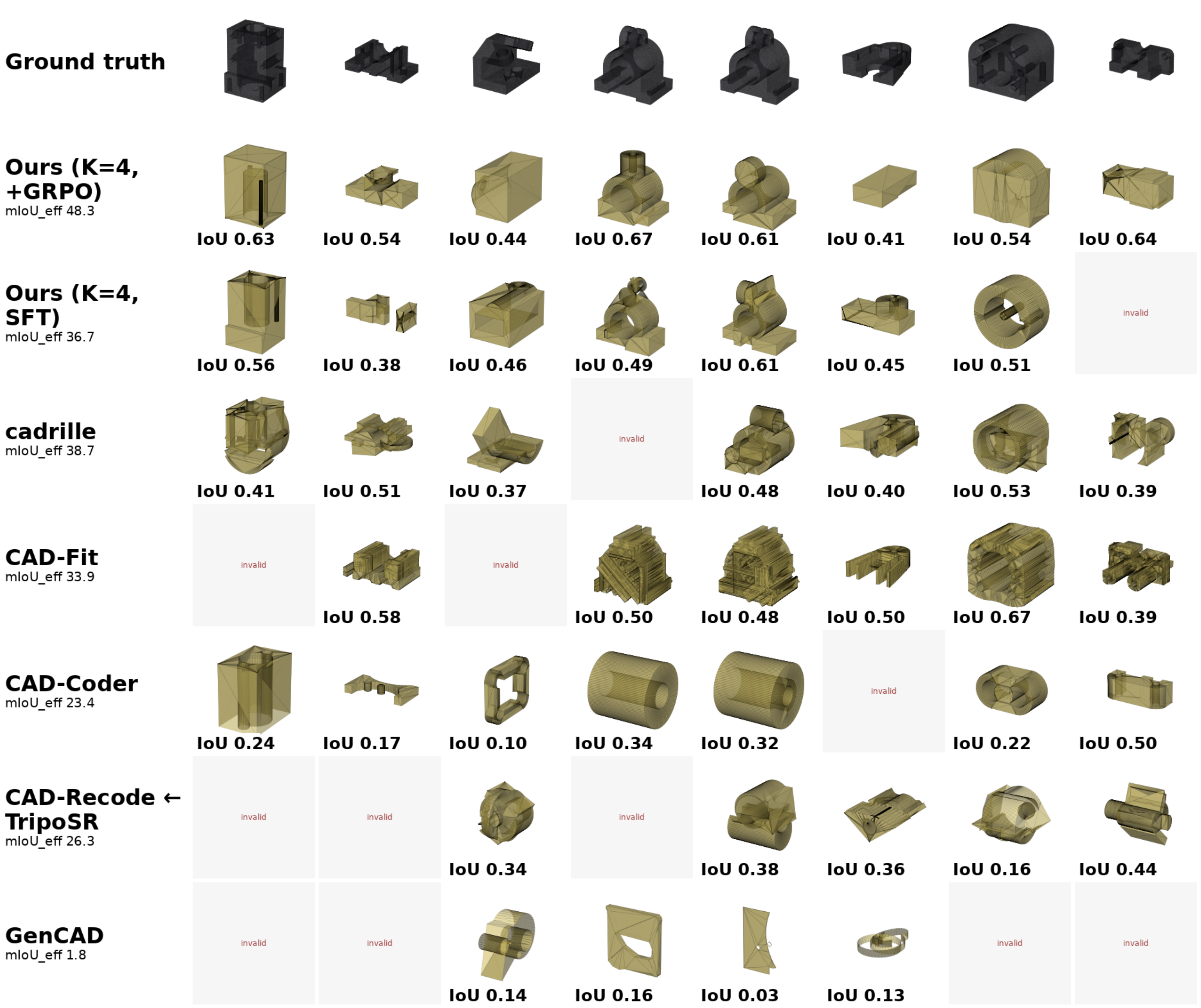}
\caption{\textbf{MP6D synth reconstructions.} Ground truth and
per-method executed solids for the objects of
Fig.~\ref{fig:qual-inputs-mp6dsynth-v2}, translucent and aligned by each
object's best-fit rotation with the corresponding IoU.}
\label{fig:qual-preds-mp6dsynth-v2}
\end{figure}

\begin{figure}[h!]
\centering
\includegraphics[width=\linewidth]{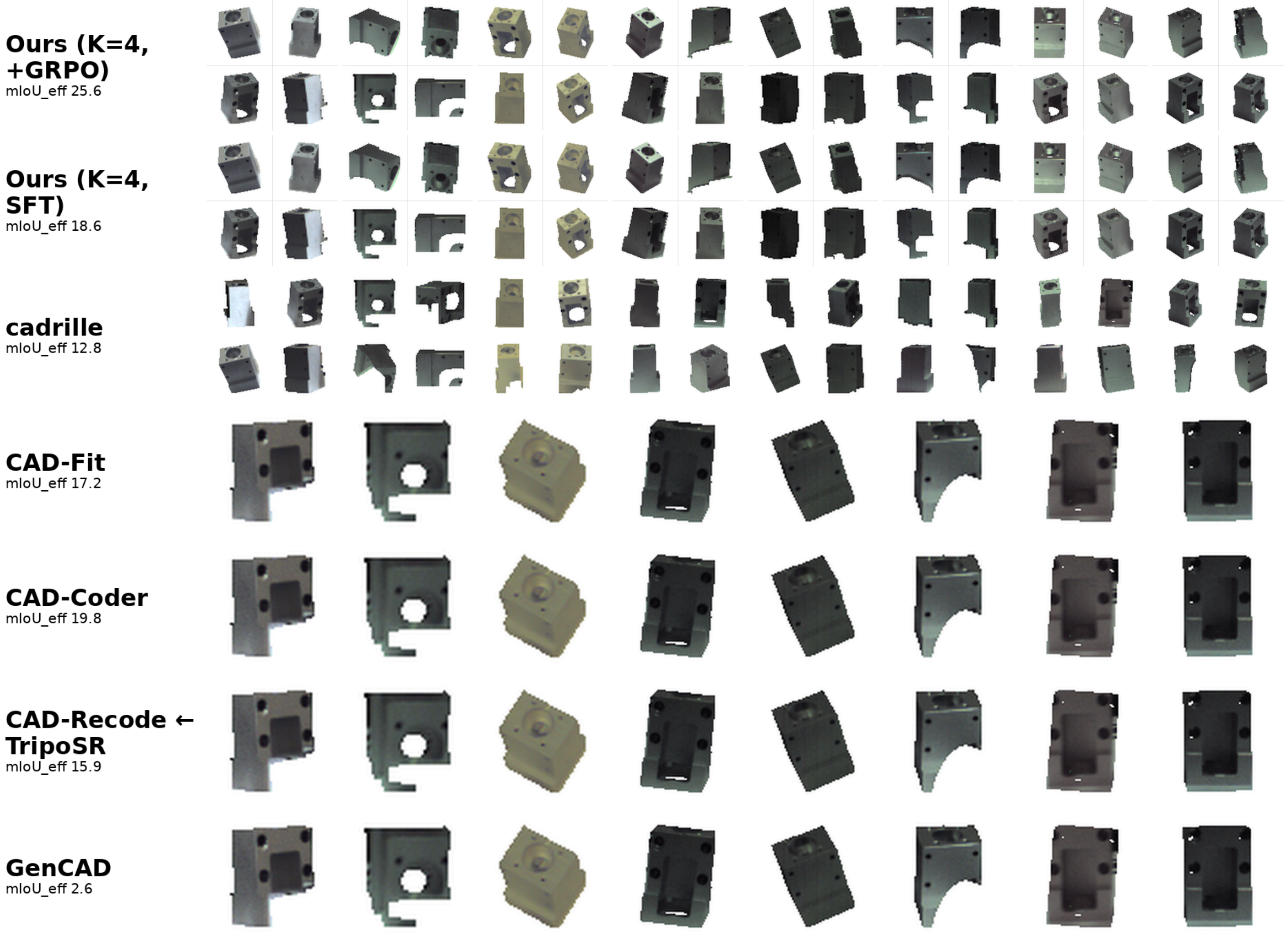}
\caption{\textbf{MP6D real model inputs.} The real
photographs fed to each method for eight MP6D captures, without any
test-time domain match (conventions as in
Fig.~\ref{fig:qual-inputs-deepcad-v2}); ours consumes the real multi-view input
directly.}
\label{fig:qual-inputs-mp6draw-v2}
\end{figure}

\begin{figure}[h!]
\centering
\includegraphics[width=\linewidth]{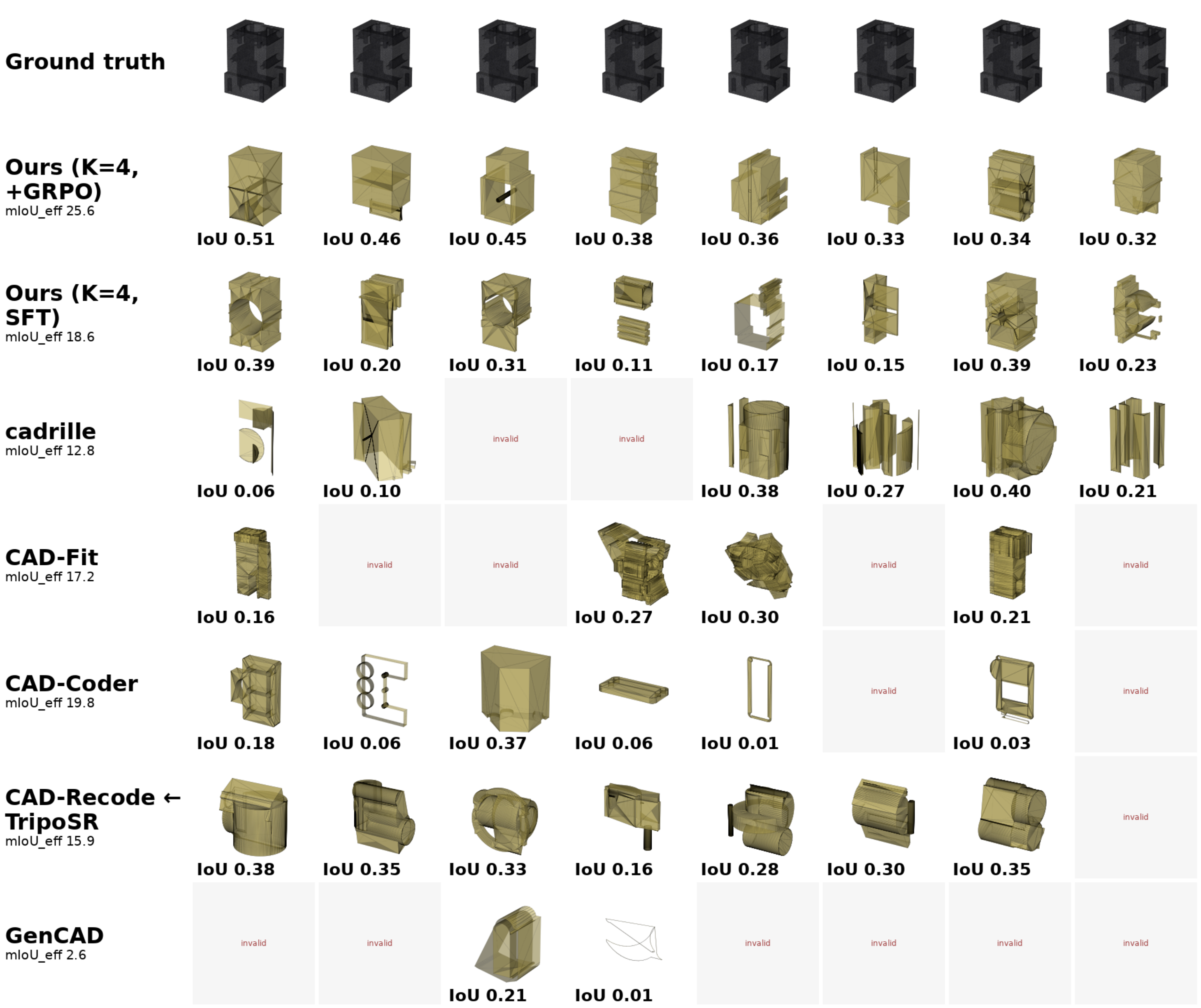}
\caption{\textbf{MP6D real reconstructions.} Ground truth
and per-method executed solids for the objects of
Fig.~\ref{fig:qual-inputs-mp6draw-v2}, translucent and aligned by each
object's best-fit rotation with the corresponding IoU. Unlike real T-LESS,
MP6D captures are acquired on a plain background, and our framework leads
every baseline on this real condition (Table~\ref{tab:mp6d}).}
\label{fig:qual-preds-mp6draw-v2}
\end{figure}

We complement the aggregate metrics with per-object qualitative comparisons on DeepCAD, Fusion360, all three T-LESS conditions (synthetic, real-raw, and real-canon), and both MP6D conditions (synthetic and real). For each benchmark we show two figures over the same set of
held-out objects: the \emph{inputs} actually consumed by every method (Fig.~\ref{fig:qual-inputs-deepcad-v2},\ref{fig:qual-inputs-fusion360-v2}, \ref{fig:qual-inputs-tlesscanon-v2}, \ref{fig:qual-inputs-tlessraw-v2}, \ref{fig:qual-inputs-tlesssynth-v2}, \ref{fig:qual-inputs-mp6dsynth-v2}, \ref{fig:qual-inputs-mp6draw-v2}), and the
executed \emph{reconstructions} they produce (Fig.~\ref{fig:qual-preds-deepcad-v2}, \ref{fig:qual-preds-fusion360-v2}, \ref{fig:qual-preds-tlesscanon-v2}, \ref{fig:qual-preds-tlessraw-v2}, \ref{fig:qual-preds-tlesssynth-v2}, \ref{fig:qual-preds-mp6dsynth-v2}, \ref{fig:qual-preds-mp6draw-v2}). To keep the comparison fair, each
baseline is fed its own native input distribution and the scene contributes \emph{only occlusion}: the single-view lift methods (CAD-Fit, CAD-Recode$\leftarrow$TripoSR) receive a clean white-background studio render from the same feasible viewpoint, cadrille its native four-quadrant view grid with scene-occluded quadrants blanked, and CAD-Coder / GenCAD their native isometric renders; only our method consumes the real scene-embedded multi-view input.
This isolates reconstruction ability from any input-domain penalty. Consistent with the quantitative
tables, our reconstructions track the ground-truth geometry most faithfully on DeepCAD, Fusion360, synthetic and canonicalized T-LESS, and synthetic and real on MP6D while the lift baselines recover coarse shape but distort feature dimensions; the sole exception is real-raw T-LESS, where the raw-photograph domain gap (removed by canonicalization) leaves our method behind the mesh-lift methods.

\clearpage
\subsection{View Selector Training and Alignment}
\label{sec:app:selector}

This section details the two auxiliary perception modules of \textbf{SightCAD}: the
view selector $\pi_{\psi}$ (architecture and SA-max warm-starting,
Sec.~\ref{sec:method:training}) and the canonical alignment network
(\emph{align-net}) used to recover the object's yaw frame on T-LESS and MP6D,
where no canonical pose is available. It also specifies the view-selection
protocol used on the real and canonicalized captures of these two benchmarks
(Sec.~\ref{sec:app:realview-selection}), where the available views do not lie
on the discrete view grid the selector was trained on.

\subsubsection{View selector architecture}
\label{sec:app:selector-arch}

The selector (Fig. \ref{fig:sa-max}) maps the full candidate view set to per-view logits
$z \in \mathbb{R}^{N}$ in four stages:

\begin{enumerate}
  \item \textbf{Per-view encoding.} Each of the $N{=}36$ candidate renders is
  encoded by a \emph{frozen} DINOv2 ViT-S/14 backbone~\citep{oquab2023dinov2};
  the CLS token gives a 384-d per-view descriptor, which a
  LayerNorm{+}Linear projection maps to the working width $d{=}256$.
  \item \textbf{Pose injection.} Each view's camera pose is added as a
  sinusoidal positional encoding. Azimuth $\theta$ is embedded circularly as
  $[\sin k\theta, \cos k\theta]_{k=1..32}$ followed by a learned linear
  projection to $d$; elevation is embedded analogously. The circular encoding
  respects the wrap-around of the azimuth ring ($360^{\circ}{=}0^{\circ}$)
  and makes angular distances explicit, so the network knows \emph{where}
  each view sits on the sphere, not just what it shows.
  \item \textbf{Cross-view reasoning.} A lightweight pre-norm transformer
  encoder (2 layers, 4 heads, FFN width $4d$, dropout $0.1$) exchanges
  information across all $N$ candidates. This stage is essential for
  \emph{subset} selection: after cross-attention, each view's representation
  reflects what the other views already cover, so the logits encode marginal
  value (complementarity, redundancy) rather than isolated per-view quality.
  \item \textbf{Scoring.} A two-layer MLP head
  (LayerNorm $\rightarrow$ Linear $256{\to}128$ $\rightarrow$ GELU
  $\rightarrow$ Linear $128{\to}1$) emits one unnormalized logit per view.
  Training samples subsets without replacement via the sequential masked
  softmax of Eq.\ref{eq:pl}; inference takes the deterministic
  top-$\Ksel$.
\end{enumerate}

With the backbone frozen, the trainable head is only ${\sim}2$M parameters,
so the selector trains on a single GPU and adds negligible inference cost on
top of the generator. Views are fed at $126 \times 126$ (a multiple of the
ViT patch size 14, avoiding internal resizing).

\subsubsection{Warm-starting the selector by SA-max distillation}
\label{sec:app:selector-warmstart}

\begin{figure}[h!]
    \centering
    \includegraphics[width=0.8\linewidth]{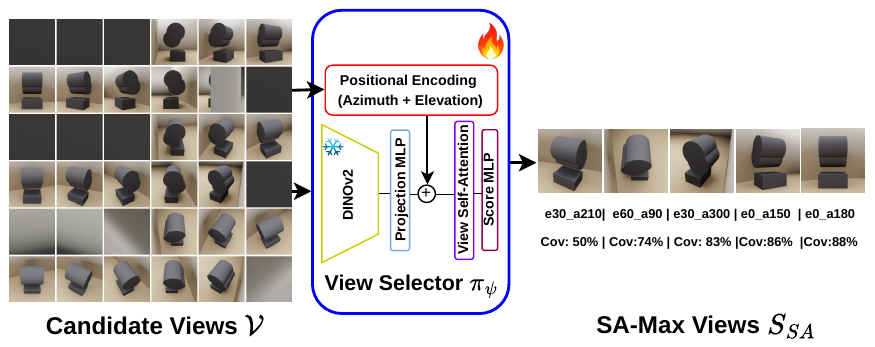}
    \caption{View-selector pretraining by SA-max: the greedy
    surface-area set-cover SA-max is distilled into an image-based scorer. The coverage (cov.) values denote the cumulative coverage of the object as the views increase from left to right. After training a central view selector we then train seperate view selectors for \emph{Joint-training} for every view budget K=$\{1,..5\}$ .}
    \label{fig:sa-max}
\end{figure}

Starting selector RL from random logits gives near-uninformative group
rewards, so we first distill the SA-max policy of Sec.~\ref{sec:data} into
the image-based scorer (Fig.~\ref{fig:sa-max}). Distillation turns the
per-placement geometric ``lookup table'' (the greedy chain stored with each
placement's visibility matrix) into a \emph{function of images and poses}
that generalizes to unseen objects and placements where no visibility oracle
exists.

\begin{enumerate}
  \item \textbf{Targets.} For each placement, the positives are the first
  five entries of the SA-greedy chain, i.e., the views that sequentially
  maximize newly visible surface area; all other views are negatives. We
  deliberately supervise the top-5 \emph{chain} rather than a binary top-4
  label: the marginal-gain gap between chain ranks \#4 and \#5 is often
  near-tied (coverage saturates around $\Ksel{=}5$), so a hard top-4 label
  would inject noise exactly at the top-$\Ksel$ decision boundary.
  \item \textbf{Set-membership loss.} A RankNet-style pairwise ranking loss
  requires every chain view to outscore every non-chain view by margin
  $m{=}1$:
  $\mathcal{L}_{\mathrm{pair}} = \operatorname{mean}_{(i,j)}
  \operatorname{softplus}\!\big(m - (z_i - z_j)\big)$
  over all (chain $i$, non-chain $j$) pairs. Unlike a multi-label BCE, no
  gradient is spent on absolute logit calibration; only the decision
  boundary that top-$\Ksel$ selection actually uses is optimized.
  \item \textbf{Within-chain order loss.} A Plackett--Luce term adds the NLL
  of the greedy chain \emph{order} under the logits (the same sequential
  masked softmax as Eq.\ref{eq:pl}), with weight $0.5$. This supervises the
  within-chain ranking, including the \#4-versus-\#5 boundary, through a
  ranking cue instead of a fabricated binary one.
  \item \textbf{Optimization.} AdamW, learning rate $10^{-4}$ with cosine
  decay, weight decay $0.01$, batch of 16 objects (each contributing all
  $N$ candidate views), trained on the full training split with the same
  train/val split seed as the generator.
\end{enumerate}

On the held-out validation split the distilled scorer reaches $55.7\%$
top-4 overlap with SA-max's exact picks and $87.6\%$ of SA-max's
union surface coverage (a tie-agnostic measure of selection quality),
against $71.0\%$ coverage for a random-4 baseline. Exact-set agreement is
low ($4.7\%$) by construction: many SA-max picks are near-ties, so the
strict-set metric penalizes selections that are equally good in coverage.
This warm-started policy is the initialization for the joint GRPO phase of
Sec.~\ref{sec:method:training}; one selector is trained per budget
$\Ksel$, which is why the per-budget optima of Sec.~\ref{sec:analysis} need
not be nested.

\subsubsection{Canonical alignment network for T-LESS and MP6D}
\label{sec:app:alignnet}

\begin{figure}[h!]
    \centering
    \includegraphics[width=\linewidth]{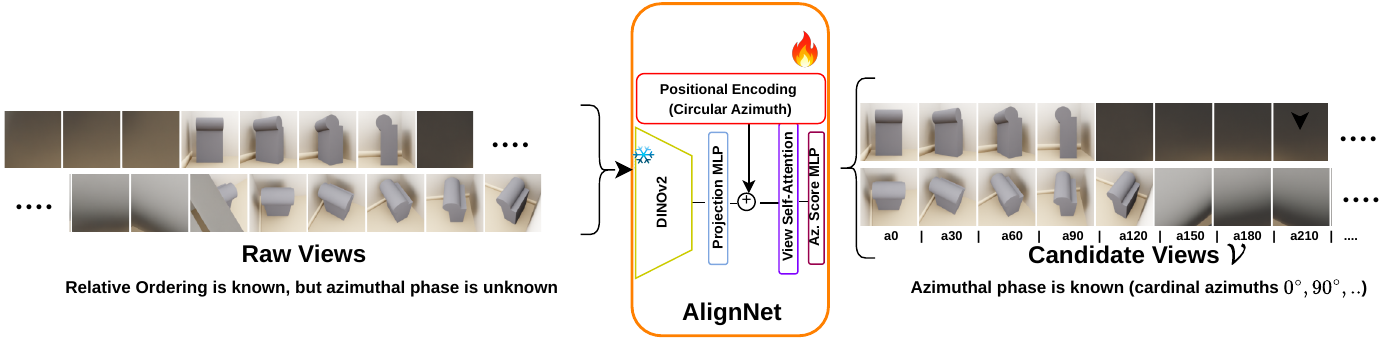}
    \caption{Align-net training for pose estimation on captured renders of CAD objects, used for the T-LESS (real and canon) and MP6D evaluations to re-anchor view azimuths to the canonical frame before CAD generation.}
    \label{fig:alignNet}
\end{figure}

On the CAD benchmarks the canonical yaw frame is known by construction; on the
industrial benchmarks T-LESS and MP6D (especially for real and canon captures
from the wild) it is not. The align-net recovers the in-plane object
orientation from images alone, so that the selected views can be re-anchored
to a canonical frame before decoding and the generator can emit a program in a
consistent frame (Fig.~\ref{fig:alignNet}).

\begin{enumerate}
  \item \textbf{Task.} Given the level ring of 12 views (elevation
  $0^{\circ}$, azimuths $30^{\circ}$ apart) presented in azimuth order with
  \emph{unknown global phase}, predict which ring position is the canonical
  $a{=}0^{\circ}$ view. This is a 12-way classification with chance
  $1/12 \approx 8.3\%$.
  \item \textbf{Architecture.} The network mirrors the view selector:
  frozen DINOv2 ViT-S/14 CLS features $\rightarrow$ projection to
  $d{=}256$ $\rightarrow$ circular azimuth encoding of the
  \emph{presentation} azimuths $0^{\circ},30^{\circ},\dots,330^{\circ}$
  (relative ring geometry is known; the absolute phase is not, so the
  encoding leaks no answer) $\rightarrow$ 2-layer cross-view transformer
  $\rightarrow$ per-position logit. Cross-view attention supplies the
  ``which face is the front'' comparison across the ring.
  \item \textbf{Training.} Frozen DINO features for each object's ring are
  cached once; the head is trained with cross-entropy on random cyclic
  rolls of the ring (the target is the rolled position of the true
  $a{=}0$ view), so every gradient step sees a fresh unknown phase. AdamW,
  learning rate $3{\times}10^{-4}$, weight decay $10^{-4}$, batch 128,
  80 epochs over 40k training objects, views at $224\times224$.
  \item \textbf{Accuracy.} On the held-out validation split, evaluated over
  all 12 rolls per object: exact top-1 $28.2\%$ (3.4$\times$ chance),
  correct up to a $180^{\circ}$ flip $51.9\%$, and correct up to a
  $90^{\circ}$ rotation $90.3\%$. The mean circular error
  ($86^{\circ}$) is dominated by these symmetry flips: most CAD parts are
  near-2- or 4-fold symmetric, making the exact anchor ill-posed while the
  \emph{axis-aligned frame} (frame modulo $90^{\circ}$) is recovered for
  $9$ in $10$ objects, which is the quantity that matters for
  sketch-and-extrude programs authored in axis-aligned workplanes.
  \item \textbf{Deployment on T-LESS and MP6D.} For each capture, the 12
  available views nearest to the $30^{\circ}$-spaced ring slots are assembled
  into a presentation ring; the align-net predicts the canonical slot; all
  view azimuths are re-anchored by the predicted offset (predicted
  $a0 \rightarrow 0^{\circ}$) before selection and decoding. The generated
  program then lives in the aligned frame. This step is applied identically
  to every condition of both benchmarks: T-LESS \emph{synth}, \emph{real},
  and \emph{canon}, and MP6D \emph{synth} and \emph{real}.
\end{enumerate}

\subsubsection{View selection on real and canonicalized multi-view captures}
\label{sec:app:realview-selection}

The view selector is trained on the fixed three-ring view lattice of
Sec.~\ref{sec:data}, but the real T-LESS and MP6D captures lie on scattered
camera poses: T-LESS provides images on discrete elevation rings whose
azimuths do not coincide with our grid, and MP6D frames come from hand-held
video sweeps whose azimuth \emph{and} elevation vary quasi-continuously
(camera poses are read off the ground-truth object-pose annotations, exactly
as in the BOP conventions). Rather than fine-tune the selector per dataset,
we reconstruct its training-time input at test time:

\begin{enumerate}
  \item \textbf{Azimuth re-anchoring.} All view azimuths are first
  re-anchored to the align-net's predicted canonical frame
  (Sec.~\ref{sec:app:alignnet}), so lattice slots carry the same meaning as
  during training.
  \item \textbf{Snap-to-lattice.} Within each elevation ring, every lattice
  slot is assigned the nearest available native view by azimuth distance,
  greedily by angular error, with at most one native view per slot and a
  snap tolerance of $20^{\circ}$; a view can occupy only one slot.
  \item \textbf{Blank-tile padding.} Lattice slots left unfilled receive a
  blank white tile, the exact appearance that scene-infeasible views have
  during training, so the reconstructed lattice is in-distribution for the
  selector and empty slots are naturally down-scored.
  \item \textbf{Selection and map-back.} The frozen selector scores the
  reconstructed lattice and the deterministic top-$\Ksel$ slots are taken;
  each pick maps back to the native view occupying its slot. Picks landing
  on blank slots are dropped and the selection is back-filled from the
  remaining native views, so exactly $\Ksel{=}4$ real views always reach the
  generator, tagged with their lattice poses.
\end{enumerate}

Elevation handling differs only in how rings are populated: T-LESS captures
(and the MP6D synthetic re-renders) carry discrete elevation rings that map
directly onto the lattice's three rings and are therefore retained, whereas
the scattered elevations of the MP6D real video frames provide no stable ring
structure, so selection there is anchored on the azimuthal lattice alone. The
entire protocol, alignment, snapping, padding, and map-back, is identical
across the synthetic, raw, and canonicalized conditions of each benchmark;
the canonicalized T-LESS condition changes only the pixels inside each view
(segmentation, exposure normalization, native-plate compositing,
Sec.~\ref{sec:experiments}), never the set of candidate views or the
selection rule.

\subsection{Dataset and Training Details}
\label{sec:app:details}

\paragraph{Wall and corner anchor curation.}
Anchor pools are obtained by scanning each room's floor on a
$0.25\,\mathrm{m}$ grid, ray-casting for perpendicular wall pairs, and
collision-checking both the object and its camera orbit against the room
geometry.

\paragraph{Uniformly spaced view policy.}
The uniform policy is a naive orbit scan: it picks one of the three elevation
rings ($0^{\circ}/30^{\circ}/60^{\circ}$) and a random azimuth phase, then
takes $K$ azimuths equispaced around that ring ($90^{\circ}$ apart at
$K{=}4$). It ignores the scene, so occluded views may be included, subject
only to a guarantee that at least one selected view sees the object.

\paragraph{Feature-weighted SA-max (fw-SA).}
fw-SA keeps SA-max's greedy set-cover chain (Sec.~\ref{sec:data}) but
reweights the visible area of each cached surface sample point by a
curvature-saliency factor $w = 0.25 + 0.75\,s$, where $s$ integrates
$|\text{dihedral angle}|\times\text{edge length}$ over the mesh's
face-adjacency edges within a radius of $3\%$ of the part diagonal around the
point, normalized by its mesh-wide mean and capped at the 95th percentile.
Each greedy step therefore appends the view adding the most newly visible
\emph{saliency-weighted} area, biasing selection toward feature-dense regions
(holes, fillets, slots) rather than blank faces. The reweighting is
consequential at the selection level, roughly two thirds of objects receive a
different view set than SA-max at every budget, while sacrificing only
$0.3$--$2.3$ pp of real surface coverage (Table~\ref{tab:policy}); its effect
on reconstruction accuracy is reported in Table~\ref{tab:schemes-main}.

\begin{table}[h]
\centering
\caption{Optimization hyperparameters of the two RL training phases
(Sec.~\ref{sec:method:training}).}
\label{tab:hparams}
\small
\setlength{\tabcolsep}{6pt}
\resizebox{\linewidth}{!}{%
\begin{tabular}{@{}l l l@{}}
\toprule
 & Joint training & Generator RL tuning \\
\midrule
Group size $G$ & 4 & 8 \\
Sampling temperature & $1.0$ (view subsets) & $0.9$ (decodes) \\
Batch size & 64 global (4 GPUs) & 12 per GPU (4 GPUs) \\
Selector optimizer & AdamW, lr $10^{-4}$ & frozen \\
Generator optimizer & AdamW, lr $10^{-5}$, wd $10^{-2}$ & AdamW, lr $10^{-5}$ \\
SFT weight & $0.5$ & --- \\
KL coefficient $\beta$ & $0$ & $0.02$ \\
Invalid-code penalty & $-0.5$ & $-0.5$ \\
SFT gate coverage floors (per-view / cumulative) & $0.05$ / $0.3$ & --- \\
\bottomrule
\end{tabular}%
}
\end{table}

\subsection{Additional Evaluations}
\label{sec:app:insitu4k}

This section completes the main-paper comparisons on the full held-out
scene-embedded test split ($n{=}4000$ CAD-Recode + Text2CAD objects,
Sec.~\ref{sec:experiments}): first the co-trained view-policy comparison at
$\Ksel{=}3$ promised in Sec.~\ref{sec:experiments}, then external CAD
baselines evaluated on the same 4k objects.

\paragraph{Every policy with its own co-trained generator ($\Ksel{=}3$).}
Table~\ref{tab:k3-cotrained} justifies the choice of SA-max as the co-trained
heuristic anchor in Table~\ref{tab:schemes-main}a (fw-SA, its feature-weighted
variant, matches SA-max there without a dedicated co-trained generator and here with a co-trained generator): at $\Ksel{=}3$, each
heuristic policy is paired with a generator fine-tuned for one epoch on
\emph{that policy's} views, starting from the same policy-neutral
initialization (Sec.~\ref{sec:method:training}); the unadapted policy-neutral
generator decoding unrestricted random views (Table~\ref{tab:schemes-main}b)
marks the no-selection floor. The policy ordering of the policy-neutral
panel is preserved under co-training, learned $>$ SA-max $\approx$ fw-SA $>$ uniform, so
SA-max remains the strongest non-learned policy even when every alternative
is given its own adapted generator. The co-adaptation gain is also unevenly
distributed: uniform improves by $+3.2$ over its policy-neutral value while
SA-max gains $+1.7$ and the learned selector $+4.2$, whose consistent,
informative view distribution gives the generator a stable input regime it
can specialize to.

\begin{table}[h]
\centering
\caption{\textbf{View policies with co-trained generators at $\Ksel{=}3$.}
Each heuristic policy's generator is fine-tuned for one epoch on that
policy's views from the same policy-neutral initialization.
Scene-embedded 4k test split, mIoU\textsubscript{eff} (\%), standard errors
$0.47$--$0.53$.}
\label{tab:k3-cotrained}
\small
\setlength{\tabcolsep}{6pt}
\begin{tabular}{@{}l cc@{}}
\toprule
Policy & mIoU\textsubscript{eff} $\uparrow$ & IR\% $\downarrow$ \\
\midrule
Random           & 36.89 & 2.03 \\
Uniform          & 43.80 & 1.85 \\
SA-max & 47.75 & 2.00 \\
fw-SA & 47.76 & 2.00 \\
Learned (Joint-Training)          & \textbf{51.69} & \textbf{1.80} \\
\bottomrule
\end{tabular}
\end{table}

\paragraph{Statistical significance and occlusion-regime breakdown.}
Table~\ref{tab:sig-regime} reports paired per-object learned$-$SA-max
contrasts on the same 4000 scene-embedded test objects, for the
policy-neutral generator (Table~\ref{tab:schemes-main}b) and the co-adapted
+GRPO arms (Table~\ref{tab:schemes-main}a). Every per-budget gap is
significant under a two-sided Wilcoxon signed-rank test: $p<0.02$ at every
$\Ksel$ for the policy-neutral decoder and $p<10^{-21}$ for the co-adapted
arms. Binning the co-adapted contrast by placement regime shows the advantage
is significant \emph{within} every regime at every budget ($p<10^{-4}$ in all
15 cells) and is at least as large on mildly occluded floor placements as in
corners.

\begin{table}[h]
\centering
\caption{\textbf{Paired learned$-$SA-max contrasts on the 4k scene-embedded
test split.} $\Delta$mIoU\textsubscript{eff} (pp), paired per object, with
task-bootstrap 95\% CIs; all entries are significant under two-sided Wilcoxon
signed-rank tests ($p<0.02$ for the policy-neutral row, $p<10^{-4}$ for the
co-adapted row and every regime cell). Regimes: 1305 floor / 933 wall /
1762 corner.}
\label{tab:sig-regime}
\small
\setlength{\tabcolsep}{4pt}
\begin{tabular}{@{}l ccccc@{}}
\toprule
$\Delta$mIoU\textsubscript{eff} (pp) & $\Ksel{=}1$ & $\Ksel{=}2$ & $\Ksel{=}3$ & $\Ksel{=}4$ & $\Ksel{=}5$ \\
\midrule
Policy-neutral & $+3.2$ {\scriptsize[2.4, 3.8]} & $+1.6$ {\scriptsize[0.8, 2.4]} & $+1.4$ {\scriptsize[0.7, 2.2]} & $+1.5$ {\scriptsize[0.7, 2.3]} & $+2.9$ {\scriptsize[2.0, 3.7]} \\
Co-adapted +GRPO & $+2.8$ {\scriptsize[2.3, 3.3]} & $+2.7$ {\scriptsize[2.1, 3.2]} & $+3.8$ {\scriptsize[3.2, 4.4]} & $+5.4$ {\scriptsize[4.8, 6.0]} & $+6.4$ {\scriptsize[5.8, 7.0]} \\
\quad floor  & $+3.8$ & $+2.8$ & $+4.1$ & $+6.0$ & $+7.3$ \\
\quad wall   & $+2.2$ & $+2.2$ & $+4.2$ & $+4.7$ & $+7.6$ \\
\quad corner & $+2.4$ & $+2.8$ & $+3.4$ & $+5.3$ & $+5.1$ \\
\bottomrule
\end{tabular}
\end{table}

\paragraph{External baselines on the full 4k split.}
Table~\ref{tab:insitu4k-baselines} reports external baselines on the same
4k scene-embedded split, with each method receiving its \emph{native} input
format and the scene contributing only occlusion (conventions as in
Sec.~\ref{sec:experiments}). \textbf{SightCAD} ($\Ksel{=}4$, +GRPO) leads every
baseline while emitting near-universally executable code.

\begin{table}[h]
\centering
\caption{\textbf{External baselines on the full scene-embedded 4k test split
($n{=}4000$ CAD-Recode + Text2CAD objects).} Every method receives its
\emph{native} input format and the scene contributes only occlusion: Our method leads all other baselines.}
\label{tab:insitu4k-baselines}
\small
\setlength{\tabcolsep}{6pt}
\begin{tabular}{@{}l rrrr@{}}
\toprule
Method & IR\% $\downarrow$ & medCD $\downarrow$ & mIoU\% $\uparrow$ & mIoU\textsubscript{eff} $\uparrow$ \\
\midrule
Ours ($\Ksel{=}4$, +GRPO)     & \textbf{0.30} & \textbf{8.10} & \textbf{66.70} & \textbf{66.50} \\
Ours ($\Ksel{=}4$, SFT)       & 1.25  & 11.68 & 60.00 & 59.25 \\
cadrille          & 17.12 & 13.02 & 54.30 & 45.01 \\
CAD-Coder         & 33.40 & 48.20 & 37.75 & 25.14 \\
CAD-Recode$\leftarrow$TripoSR & 43.23 & 84.94 & 18.43 & 10.46 \\
GenCAD            & 77.00 & 73.18 & 18.72 & 4.31  \\
\bottomrule
\end{tabular}
\end{table}

\end{document}

%% file: math_commands.tex

\usepackage{amsmath,amsfonts,bm}









\def\eqref#1{equation~\ref{#1}}









\def\1{\bm{1}}










\DeclareMathAlphabet{\mathsfit}{\encodingdefault}{\sfdefault}{m}{sl}
\SetMathAlphabet{\mathsfit}{bold}{\encodingdefault}{\sfdefault}{bx}{n}













%% file: main_arxiv.bbl
\begin{thebibliography}{49}
\providecommand{\natexlab}[1]{#1}
\providecommand{\url}[1]{\texttt{#1}}
\expandafter\ifx\csname urlstyle\endcsname\relax
  \providecommand{\doi}[1]{doi: #1}\else
  \providecommand{\doi}{doi: \begingroup \urlstyle{rm}\Url}\fi

\bibitem[Alam \& Ahmed(2024)Alam and Ahmed]{alam2024gencad}
Md~Ferdous Alam and Faez Ahmed.
\newblock Gencad: Image-conditioned computer-aided design generation with
  transformer-based contrastive representation and diffusion priors.
\newblock \emph{arXiv preprint arXiv:2409.16294}, 2024.

\bibitem[Bai et~al.(2026)Bai, Wang, Liu, Luo, Wen, Dai, Chen, Chen, Liu, Li,
  et~al.]{bai2026learning}
Yongjie Bai, Zhouxia Wang, Yang Liu, Kaijun Luo, Yifan Wen, Mingtong Dai,
  Weixing Chen, Ziliang Chen, Lingbo Liu, Guanbin Li, et~al.
\newblock Learning to see and act: Task-aware virtual view exploration for
  robotic manipulation.
\newblock In \emph{Proceedings of the IEEE/CVF Conference on Computer Vision
  and Pattern Recognition}, pp.\  13386--13396, 2026.

\bibitem[Bali \& Aydin(2026)Bali and Aydin]{bali2026mv}
Kartik Bali and Roland Aydin.
\newblock Mv-gel: Language-driven multi-view geometric entity localization on
  meshes.
\newblock In \emph{European Conference on Computer Vision}, pp.\  599--617.
  Springer, 2026.

\bibitem[Chen et~al.(2022)Chen, Yang, Wu, and Wu]{chen2022mp6d}
Long Chen, Han Yang, Chenrui Wu, and Shiqing Wu.
\newblock Mp6d: An rgb-d dataset for metal parts’ 6d pose estimation.
\newblock \emph{IEEE Robotics and Automation Letters}, 7\penalty0 (3):\penalty0
  5912--5919, 2022.

\bibitem[Chen et~al.(2024)Chen, Li, Wang, Xue, and Pang]{chen2024gennbv}
Xiao Chen, Quanyi Li, Tai Wang, Tianfan Xue, and Jiangmiao Pang.
\newblock Gennbv: Generalizable next-best-view policy for active 3d
  reconstruction.
\newblock In \emph{2024 IEEE/CVF Conference on Computer Vision and Pattern
  Recognition (CVPR)}, pp.\  16436--16445. IEEE, 2024.

\bibitem[Connolly(1985)]{connolly1985determination}
Cl~Connolly.
\newblock The determination of next best views.
\newblock In \emph{Proceedings. 1985 IEEE international conference on robotics
  and automation}, volume~2, pp.\  432--435. IEEE, 1985.

\bibitem[Csiszar \& Narayan(1995)Csiszar and Narayan]{csiszar1995channel}
Imre Csiszar and Prakash Narayan.
\newblock Channel capacity for a given decoding metric.
\newblock \emph{IEEE Transactions on Information Theory}, 41\penalty0
  (1):\penalty0 35--43, 1995.

\bibitem[Delmerico et~al.(2018)Delmerico, Isler, Sabzevari, and
  Scaramuzza]{delmerico2018comparison}
Jeffrey Delmerico, Stefan Isler, Reza Sabzevari, and Davide Scaramuzza.
\newblock A comparison of volumetric information gain metrics for active 3d
  object reconstruction.
\newblock \emph{Autonomous Robots}, 42\penalty0 (2):\penalty0 197--208, 2018.

\bibitem[Denninger et~al.()Denninger, Winkelbauer, Sundermeyer, Strobl, Humt,
  and Triebel]{denningerblenderproc2}
M~Denninger, D~Winkelbauer, M~Sundermeyer, KH~Strobl, M~Humt, and R~Triebel.
\newblock Blenderproc2: a procedural pipeline for photorealistic rendering. j.
  open source softw. 8 (82), 4901 (2023).

\bibitem[Doris et~al.(2026)Doris, Alam, Heyrani~Nobari, and
  Ahmed]{doris2026cad}
Anna~C Doris, Ferdous Alam, Amin Heyrani~Nobari, and Faez Ahmed.
\newblock Cad-coder: An open-source vision-language model for computer-aided
  design code generation.
\newblock \emph{Journal of Mechanical Design}, 148\penalty0 (7):\penalty0
  071702, 2026.

\bibitem[Foster et~al.(2021)Foster, Ivanova, Malik, and
  Rainforth]{foster2021deep}
Adam Foster, Desi~R Ivanova, Ilyas Malik, and Tom Rainforth.
\newblock Deep adaptive design: Amortizing sequential bayesian experimental
  design.
\newblock In \emph{International conference on machine learning}, pp.\
  3384--3395. PMLR, 2021.

\bibitem[Fraser et~al.(1984)]{fraser1984network}
Clive~S Fraser et~al.
\newblock Network design considerations for non-topographic photogrammetry.
\newblock \emph{Photogrammetric Engineering and Remote Sensing}, 50\penalty0
  (8):\penalty0 1115--1126, 1984.

\bibitem[Gao et~al.(2026)Gao, Zhang, Ni, Zhu, Li, Wang, and
  Xu]{gao2026occamview}
Hongbo Gao, Wei Zhang, Zeyu Ni, Dihao Zhu, Ruifeng Li, Yunke Wang, and Chang
  Xu.
\newblock Occamview: Object-conditioned view selection for frame-budgeted
  active 3d gaussian reconstruction.
\newblock \emph{arXiv preprint arXiv:2608.16499}, 2026.

\bibitem[G{\"a}rtner et~al.(2020)G{\"a}rtner, Pirinen, and
  Sminchisescu]{gartner2020deep}
Erik G{\"a}rtner, Aleksis Pirinen, and Cristian Sminchisescu.
\newblock Deep reinforcement learning for active human pose estimation.
\newblock In \emph{Proceedings of the AAAI Conference on Artificial
  Intelligence}, volume~34, pp.\  10835--10844, 2020.

\bibitem[Handa et~al.(2016)Handa, Patraucean, Badrinarayanan, Stent, and
  Cipolla]{handa2016understanding}
Ankur Handa, Viorica Patraucean, Vijay Badrinarayanan, Simon Stent, and Roberto
  Cipolla.
\newblock Understanding real world indoor scenes with synthetic data.
\newblock In \emph{Proceedings of the IEEE conference on computer vision and
  pattern recognition}, pp.\  4077--4085, 2016.

\bibitem[Hodan et~al.(2017)Hodan, Haluza, Obdr{\v{z}}{\'a}lek, Matas, Lourakis,
  and Zabulis]{hodan2017t}
Tom{\'a}{\v{s}} Hodan, Pavel Haluza, {\v{S}}tep{\'a}n Obdr{\v{z}}{\'a}lek, Jiri
  Matas, Manolis Lourakis, and Xenophon Zabulis.
\newblock T-less: An rgb-d dataset for 6d pose estimation of texture-less
  objects.
\newblock In \emph{2017 IEEE winter conference on applications of computer
  vision (WACV)}, pp.\  880--888. IEEE, 2017.

\bibitem[Hou et~al.(2023)Hou, Gould, and Zheng]{hou2023learning}
Yunzhong Hou, Stephen Gould, and Liang Zheng.
\newblock Learning to select camera views: Efficient multiview understanding at
  few glances.
\newblock \emph{arXiv preprint arXiv:2303.06145}, 2023.

\bibitem[Isler et~al.(2016)Isler, Sabzevari, Delmerico, and
  Scaramuzza]{isler2016information}
Stefan Isler, Reza Sabzevari, Jeffrey Delmerico, and Davide Scaramuzza.
\newblock An information gain formulation for active volumetric 3d
  reconstruction.
\newblock In \emph{2016 IEEE international conference on robotics and
  automation (ICRA)}, pp.\  3477--3484. IEEE, 2016.

\bibitem[Jayaraman \& Grauman(2016)Jayaraman and Grauman]{jayaraman2016look}
Dinesh Jayaraman and Kristen Grauman.
\newblock Look-ahead before you leap: end-to-end active recognition by
  forecasting the effect of motion.
\newblock In \emph{European Conference on Computer Vision}, pp.\  489--505.
  Springer, 2016.

\bibitem[Jiang et~al.(2023)Jiang, Lei, and Daniilidis]{jiang2023fisherrf}
Wen Jiang, Boshu Lei, and Kostas Daniilidis.
\newblock Fisherrf: Active view selection and uncertainty quantification for
  radiance fields using fisher information.
\newblock \emph{arXiv preprint arXiv:2311.17874}, 2023.

\bibitem[Jin et~al.(2023)Jin, Chen, R{\"u}ckin, and Popovi{\'c}]{jin2023neu}
Liren Jin, Xieyuanli Chen, Julius R{\"u}ckin, and Marija Popovi{\'c}.
\newblock Neu-nbv: Next best view planning using uncertainty estimation in
  image-based neural rendering.
\newblock In \emph{2023 IEEE/RSJ International Conference on Intelligent Robots
  and Systems (IROS)}, pp.\  11305--11312. IEEE, 2023.

\bibitem[Khan et~al.(2024)Khan, Sinha, Sheikh, Stricker, Ali, and
  Afzal]{khan2024text2cad}
Mohammad~S Khan, Sankalp Sinha, Talha~U Sheikh, Didier Stricker, Sk~A Ali, and
  Muhammad~Z Afzal.
\newblock Text2cad: Generating sequential cad designs from beginner-to-expert
  level text prompts.
\newblock \emph{Advances in Neural Information Processing Systems},
  37:\penalty0 7552--7579, 2024.

\bibitem[Kirillov et~al.(2023)Kirillov, Mintun, Ravi, Mao, Rolland, Gustafson,
  Xiao, Whitehead, Berg, Lo, et~al.]{kirillov2023segment}
Alexander Kirillov, Eric Mintun, Nikhila Ravi, Hanzi Mao, Chloe Rolland, Laura
  Gustafson, Tete Xiao, Spencer Whitehead, Alexander~C Berg, Wan-Yen Lo, et~al.
\newblock Segment anything.
\newblock In \emph{2023 IEEE/CVF international conference on computer vision
  (ICCV)}, pp.\  3992--4003. IEEE, 2023.

\bibitem[Kolodiazhnyi et~al.(2025)Kolodiazhnyi, Tarasov, Zhemchuzhnikov,
  Nikulin, Zisman, Vorontsova, Konushin, Kurenkov, and
  Rukhovich]{kolodiazhnyi2025cadrille}
Maksim Kolodiazhnyi, Denis Tarasov, Dmitrii Zhemchuzhnikov, Alexander Nikulin,
  Ilya Zisman, Anna Vorontsova, Anton Konushin, Vladislav Kurenkov, and Danila
  Rukhovich.
\newblock cadrille: Multi-modal cad reconstruction with reinforcement learning.
\newblock \emph{arXiv preprint arXiv:2505.22914}, 2025.

\bibitem[Krause \& Golovin(2014)Krause and Golovin]{krause2014tractability}
Andreas Krause and Daniel Golovin.
\newblock Tractability: Practical approaches to hard problems.
\newblock In \emph{Submodular Function Maximization}, pp.\  71--104. Cambridge
  Univ. Press Cambridge, UK, 2014.

\bibitem[Lapidoth(1996)]{lapidoth1996mismatched}
Amos Lapidoth.
\newblock Mismatched decoding and the multiple-access channel.
\newblock \emph{IEEE Transactions on Information Theory}, 42\penalty0
  (5):\penalty0 1439--1452, 1996.

\bibitem[Lee et~al.(2022)Lee, Chen, Wang, Liniger, Kumar, and
  Yu]{lee2022uncertainty}
Soomin Lee, Le~Chen, Jiahao Wang, Alexander Liniger, Suryansh Kumar, and Fisher
  Yu.
\newblock Uncertainty guided policy for active robotic 3d reconstruction using
  neural radiance fields.
\newblock \emph{IEEE Robotics and Automation Letters}, 7\penalty0 (4):\penalty0
  12070--12077, 2022.

\bibitem[Li et~al.(2026)Li, Luo, Lou, and Zhou]{li2026recad}
Jiahao Li, Yusheng Luo, Yunzhong Lou, and Xiangdong Zhou.
\newblock Recad: Reinforcement learning enhanced parametric cad model
  generation with vision-language models.
\newblock In \emph{Proceedings of the AAAI Conference on Artificial
  Intelligence}, volume~40, pp.\  6190--6198, 2026.

\bibitem[Li et~al.(2020)Li, Zhao, Varma, Salpekar, Noordhuis, Li, Paszke,
  Smith, Vaughan, Damania, et~al.]{li2020pytorch}
Shen Li, Yanli Zhao, Rohan Varma, Omkar Salpekar, Pieter Noordhuis, Teng Li,
  Adam Paszke, Jeff Smith, Brian Vaughan, Pritam Damania, et~al.
\newblock Pytorch distributed: Experiences on accelerating data parallel
  training.
\newblock \emph{arXiv preprint arXiv:2006.15704}, 2020.

\bibitem[Lindley(1956)]{lindley1956measure}
Dennis~V Lindley.
\newblock On a measure of the information provided by an experiment.
\newblock \emph{The Annals of Mathematical Statistics}, 27\penalty0
  (4):\penalty0 986--1005, 1956.

\bibitem[Mendoza et~al.(2020)Mendoza, Vasquez-Gomez, Taud, Sucar, and
  Reta]{mendoza2020supervised}
Miguel Mendoza, J~Irving Vasquez-Gomez, Hind Taud, L~Enrique Sucar, and
  Carolina Reta.
\newblock Supervised learning of the next-best-view for 3d object
  reconstruction.
\newblock \emph{Pattern Recognition Letters}, 133:\penalty0 224--231, 2020.

\bibitem[Nehme et~al.(2026)Nehme, Whalen, and Ahmed]{nehme2026cadfit}
Ghadi Nehme, Eamon Whalen, and Faez Ahmed.
\newblock Cadfit: Precise mesh-to-cad program generation with hybrid
  optimization.
\newblock \emph{arXiv preprint arXiv:2605.01171}, 2026.

\bibitem[Nemhauser et~al.(1978)Nemhauser, Wolsey, and
  Fisher]{nemhauser1978analysis}
George~L Nemhauser, Laurence~A Wolsey, and Marshall~L Fisher.
\newblock An analysis of approximations for maximizing submodular set
  functions—i.
\newblock \emph{Mathematical programming}, 14\penalty0 (1):\penalty0 265--294,
  1978.

\bibitem[Oquab et~al.(2023)Oquab, Darcet, Moutakanni, Vo, Szafraniec, Khalidov,
  Fernandez, Haziza, Massa, El-Nouby, et~al.]{oquab2023dinov2}
Maxime Oquab, Timoth{\'e}e Darcet, Th{\'e}o Moutakanni, Huy Vo, Marc
  Szafraniec, Vasil Khalidov, Pierre Fernandez, Daniel Haziza, Francisco Massa,
  Alaaeldin El-Nouby, et~al.
\newblock Dinov2: Learning robust visual features without supervision.
\newblock \emph{arXiv preprint arXiv:2304.07193}, 2023.

\bibitem[Palmer(1981)]{palmer1981cannonical}
Stephen~E Palmer.
\newblock Cannonical perspective and the perception of objects.
\newblock \emph{Attention and performance}, 9:\penalty0 135--151, 1981.

\bibitem[Pan et~al.(2022)Pan, Lai, Song, and Huang]{pan2022activenerf}
Xuran Pan, Zihang Lai, Shiji Song, and Gao Huang.
\newblock Activenerf: Learning where to see with uncertainty estimation.
\newblock In \emph{European Conference on Computer Vision}, pp.\  230--246.
  Springer, 2022.

\bibitem[Peralta et~al.(2020)Peralta, Casimiro, Nilles, Aguilar, Atienza, and
  Cajote]{peralta2020next}
Daryl Peralta, Joel Casimiro, Aldrin~Michael Nilles, Justine~Aletta Aguilar,
  Rowel Atienza, and Rhandley Cajote.
\newblock Next-best view policy for 3d reconstruction.
\newblock In \emph{European Conference on Computer Vision}, pp.\  558--573.
  Springer, 2020.

\bibitem[Polonsky et~al.(2005)Polonsky, Patan{\'e}, Biasotti, Gotsman, and
  Spagnuolo]{polonsky2005s}
Oleg Polonsky, Giuseppe Patan{\'e}, Silvia Biasotti, Craig Gotsman, and Michela
  Spagnuolo.
\newblock What’s in an image? towards the computation of the “best” view
  of an object.
\newblock \emph{The Visual Computer}, 21\penalty0 (8):\penalty0 840--847, 2005.

\bibitem[Rukhovich et~al.(2025)Rukhovich, Dupont, Mallis, Cherenkova, Kacem,
  and Aouada]{rukhovich2025cad}
Danila Rukhovich, Elona Dupont, Dimitrios Mallis, Kseniya Cherenkova, Anis
  Kacem, and Djamila Aouada.
\newblock Cad-recode: Reverse engineering cad code from point clouds.
\newblock In \emph{2025 IEEE/CVF International Conference on Computer Vision
  (ICCV)}, pp.\  9801--9811. IEEE, 2025.

\bibitem[Scott et~al.(2003)Scott, Roth, and Rivest]{scott2003view}
William~R Scott, Gerhard Roth, and Jean-Fran{\c{c}}ois Rivest.
\newblock View planning for automated three-dimensional object reconstruction
  and inspection.
\newblock \emph{ACM Computing Surveys (CSUR)}, 35\penalty0 (1):\penalty0
  64--96, 2003.

\bibitem[Shao et~al.(2024)Shao, Wang, Zhu, Xu, Song, Bi, Zhang, Zhang, Li, Wu,
  et~al.]{shao2024deepseekmath}
Zhihong Shao, Peiyi Wang, Qihao Zhu, Runxin Xu, Junxiao Song, Xiao Bi, Haowei
  Zhang, Mingchuan Zhang, YK~Li, Yang Wu, et~al.
\newblock Deepseekmath: Pushing the limits of mathematical reasoning in open
  language models.
\newblock \emph{arXiv preprint arXiv:2402.03300}, 2024.

\bibitem[Singh \& Joachims(2019)Singh and Joachims]{singh2019policy}
Ashudeep Singh and Thorsten Joachims.
\newblock Policy learning for fairness in ranking.
\newblock \emph{Advances in neural information processing systems}, 32, 2019.

\bibitem[Song et~al.(2026)Song, Li, Li, He, Wang, and Zhang]{song2026go}
Yan Song, Zhihao Li, Chenglong Li, Li~He, Yan Wang, and Wenqiang Zhang.
\newblock Go-pre: Goal-oriented next-best-view selection via predictive
  rendering entropy for active 3d reconstruction.
\newblock \emph{arXiv preprint arXiv:2607.29037}, 2026.

\bibitem[Tochilkin et~al.(2024)Tochilkin, Pankratz, Liu, Huang, Letts, Li,
  Liang, Laforte, Jampani, and Cao]{tochilkin2024triposr}
Dmitry Tochilkin, David Pankratz, Zexiang Liu, Zixuan Huang, Adam Letts,
  Yangguang Li, Ding Liang, Christian Laforte, Varun Jampani, and Yan-Pei Cao.
\newblock Triposr: Fast 3d object reconstruction from a single image.
\newblock \emph{arXiv preprint arXiv:2403.02151}, 2024.

\bibitem[V{\'a}zquez et~al.(2002)V{\'a}zquez, Feixas, Sbert, and
  Llobet]{vazquez2002viewpoint}
Pere-Pau V{\'a}zquez, Miquel Feixas, Mateu Sbert, and Antoni Llobet.
\newblock Viewpoint entropy: a new tool for obtaining good views of molecules.
\newblock In \emph{ACM International Conference Proceeding Series}, volume~22,
  pp.\  183--188, 2002.

\bibitem[Wang et~al.(2024)Wang, Bai, Tan, Wang, Fan, Bai, Chen, Liu, Wang, Ge,
  et~al.]{wang2024qwen2}
Peng Wang, Shuai Bai, Sinan Tan, Shijie Wang, Zhihao Fan, Jinze Bai, Keqin
  Chen, Xuejing Liu, Jialin Wang, Wenbin Ge, et~al.
\newblock Qwen2-vl: Enhancing vision-language model's perception of the world
  at any resolution.
\newblock \emph{arXiv preprint arXiv:2409.12191}, 2024.

\bibitem[Wu et~al.(2021)Wu, Xiao, and Zheng]{wu2021deepcad}
Rundi Wu, Chang Xiao, and Changxi Zheng.
\newblock Deepcad: A deep generative network for computer-aided design models.
\newblock In \emph{2021 IEEE/CVF International Conference on Computer Vision
  (ICCV)}, pp.\  6752--6762. IEEE, 2021.

\bibitem[Zeng et~al.(2020)Zeng, Zhao, and Liu]{zeng2020pc}
Rui Zeng, Wang Zhao, and Yong-Jin Liu.
\newblock Pc-nbv: A point cloud based deep network for efficient next best view
  planning.
\newblock In \emph{2020 IEEE/RSJ International Conference on Intelligent Robots
  and Systems (IROS)}, pp.\  7050--7057. IEEE, 2020.

\bibitem[Zhu et~al.(2026)Zhu, Sell{\'a}n, and Terenin]{zhu2026bayesian}
Jingsen Zhu, Silvia Sell{\'a}n, and Alexander Terenin.
\newblock A bayesian approach for task-specific next-best-view selection with
  uncertain geometry.
\newblock \emph{arXiv preprint arXiv:2605.05095}, 2026.

\end{thebibliography}
